\documentclass[runningheads]{llncs}
\usepackage{graphicx}

\usepackage[dvipsnames]{xcolor}
\usepackage{tikz}
\usepackage{comment}
\usepackage{amsmath,amssymb} 
\usepackage{times}
\usepackage{epsfig}
\usepackage{graphicx}
\usepackage{amsmath}
\usepackage{amssymb}
\usepackage{enumitem}
\usepackage{booktabs}
\usepackage{multirow}
\usepackage{makecell}
\usepackage{multicol}
\usepackage{rotating}
\usepackage[T1]{fontenc}
\usepackage{pifont}  
\usepackage{xspace}
\usepackage{microtype}
\usepackage{makecell}
\usepackage{bm}  
\usepackage{bbm}  
\usepackage{fontawesome5}  
\usepackage{textcomp}  
\usepackage{gensymb}  
\usepackage{colortbl}  
\usepackage{wrapfig}

\usepackage[accsupp]{axessibility}  

\usepackage[width=122mm,left=12mm,paperwidth=146mm,height=193mm,top=12mm,paperheight=217mm]{geometry}

\usepackage[pagebackref,breaklinks,colorlinks]{hyperref}
\hypersetup{ 
    citecolor=ForestGreen,
    urlcolor=MidnightBlue,
}

\usepackage[capitalize]{cleveref}
\crefname{section}{Sec.}{Secs.}
\Crefname{section}{Section}{Sections}
\Crefname{table}{Table}{Tables}
\crefname{table}{Tab.}{Tabs.}

\newcommand*{\eg}{e.g.\@\xspace}

\newcommand*{\etal}{et~al.\@\xspace}
\newcommand*{\etc}{etc.\@\xspace}
\newcommand*{\cf}{cf.\@\xspace}

\newlength{\tabcolsepdefault}
\DeclareFontFamily{U}{mathb}{}
\DeclareFontShape{U}{mathb}{m}{n}{
  <-5.5> mathb5
  <5.5-6.5> mathb6
  <6.5-7.5> mathb7
  <7.5-8.5> mathb8
  <8.5-9.5> mathb9
  <9.5-11.5> mathb10
  <11.5-> mathb12
}{}
\DeclareSymbolFont{mathb}{U}{mathb}{m}{n}
\DeclareMathSymbol{\drsh}{3}{mathb}{"EB}

\DeclareMathOperator*{\argmax}{arg\,max}
\DeclareMathOperator*{\argmin}{arg\,min}

\newcommand{\PAR}[1]{\vskip4pt \noindent{\bf #1~}}

\renewcommand{\b}[1]{\textbf{#1}}
\definecolor{Red}{RGB}{240,0,0}
\definecolor{Green}{RGB}{0,180,0}
\newcommand{\red}[1]{\textcolor{Red}{#1}}
\newcommand{\green}[1]{\textcolor{Green}{#1}}

\newcommand{\emoji}[1]{\includegraphics[height=9pt]{#1}}
\newcommand{\iconNight}{\emoji{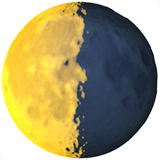}}%
\newcommand{\iconStruct}{\emoji{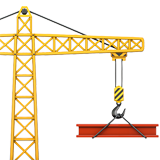}}%
\newcommand{\iconWeather}{\emoji{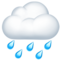}}%
\newcommand{\iconChair}{\emoji{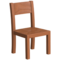}}%
\newcommand{\iconDynamic}{\emoji{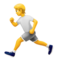}}%
\newcommand{\iconX}{\emoji{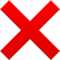}}%
\newcommand{\iconV}{\emoji{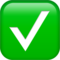}}%
\newcommand{\cmark}{\iconV}%
\newcommand{\xmark}{\iconX}%
\newcommand{\scmark}{\ding{51}}%
\newcommand{\sxmark}{\ding{55}}%

\newcommand*\samethanks[1][\value{footnote}]{\footnotemark[#1]}

\setlist[itemize]{noitemsep, topsep=0pt}
\setlist[enumerate]{noitemsep, topsep=0pt}

\newif\ifsupponly
\supponlyfalse
\newif\ifincludesupp
\includesupptrue

\begin{document}
\pagestyle{headings}
\mainmatter
\def\ECCVSubNumber{930}  

\newcommand{\DATASET}{LaMAR}
\title{\DATASET{}: Benchmarking Localization and Mapping\texorpdfstring{\\}{}for Augmented Reality} 

%
\author{Paul-Edouard Sarlin\thanks{Equal contribution. \ensuremath{{}^\dagger}\ Now at Lund University, Sweden.}\inst{1} \and
Mihai Dusmanu\samethanks\inst{1} \and
Johannes L.~Sch\"onberger\inst{2} \and
Pablo Speciale\inst{2} \and
Lukas Gruber\inst{2} \and
Viktor Larsson\ensuremath{{}^\dagger}\inst{1} \and
Ondrej Miksik\inst{2} \and
Marc Pollefeys\inst{1,2}}
\authorrunning{Sarlin and Dusmanu \etal}
%
\institute{Department of Computer Science, ETH Zurich, Switzerland \and
Microsoft Mixed Reality \& AI Lab, Zurich, Switzerland}

\ifsupponly\else 
\maketitle

\begin{figure}[!h]
    \centering
    \includegraphics[width=\textwidth]{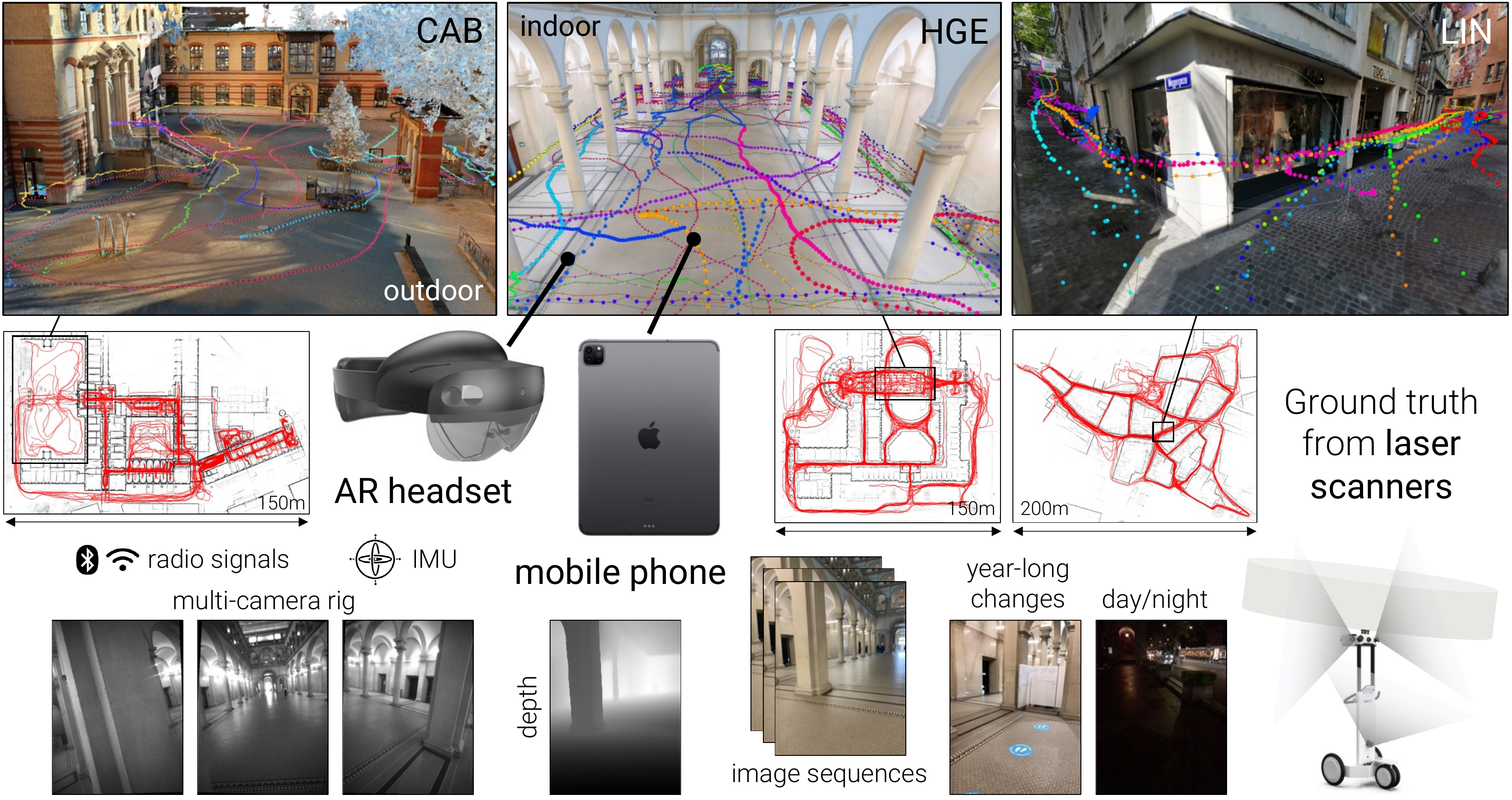}%
    \caption{%
    We revisit localization and mapping in the context of Augmented Reality by introducing LaMAR, a large-scale dataset captured using AR devices (HoloLens 2, iPhone) and laser scanners.
    }%
    \label{fig:teaser}
    \vspace{-8mm}
\end{figure}

\begin{abstract}
Localization and mapping is the foundational technology for augmented reality (AR) that enables sharing and persistence of digital content in the real world.
While significant progress has been made, researchers are still mostly driven by unrealistic benchmarks not representative of real-world AR scenarios.
These benchmarks are often based on small-scale datasets with low scene diversity, captured from stationary cameras, and lack other sensor inputs like inertial, radio, or depth data.
Furthermore, their ground-truth (GT) accuracy is mostly insufficient to satisfy AR requirements.
To close this gap, we introduce LaMAR, a new benchmark with a comprehensive capture and GT pipeline that co-registers realistic trajectories and sensor streams captured by heterogeneous AR devices in large, unconstrained scenes.
To establish an accurate GT, our pipeline robustly aligns the trajectories against laser scans in a fully automated manner.
As a result, we publish a benchmark dataset of diverse and large-scale scenes recorded with head-mounted and hand-held AR devices.
We extend several state-of-the-art methods to take advantage of the AR-specific setup and evaluate them on our benchmark.
The results offer new insights on current research and reveal promising avenues for future work in the field of localization and mapping for AR.
\end{abstract}

\section{Introduction}

Placing virtual content in the physical 3D world, persisting it over time, and sharing it with other users are typical scenarios for Augmented Reality (AR).
In order to reliably overlay virtual content in the real world with pixel-level precision, these scenarios require AR devices to accurately determine their 6-DoF pose at any point in time.
While visual localization and mapping is one of the most studied problems in computer vision, its use for AR entails specific challenges and opportunities.
First, modern AR devices, such as mobile phones or the Microsoft HoloLens or \mbox{MagicLeap One}, are often equipped with multiple cameras and additional inertial or radio sensors.
Second, they exhibit characteristic hand-held or head-mounted motion patterns.
The on-device real-time tracking systems provide spatially-posed sensor streams. 
However, many AR scenarios require positioning beyond local tracking, both indoors and outdoors, and robustness to common temporal changes of appearance and structure.
Furthermore, given the plurality of temporal sensor data, the question is often not whether, but how quickly can the device localize at any time to ensure a compelling end-user experience.
Finally, as AR adoption grows, crowd-sourced data captured by users with diverse devices can be mined for building large-scale maps without a manual and costly scanning effort.
Crowd-sourcing offers great opportunities but poses additional challenges on the robustness of algorithms, \eg, to enable cross-device localization~\cite{dusmanu2021cross}, mapping from incomplete data with low accuracy~\cite{Schoenberger2016Structure,brachmann2021limits}, privacy-preservation of data~\cite{speciale2019a,geppert2020privacy,shibuya2020privacy,geppert2021privacy,dusmanu2021privacy}, \etc

However, the academic community is mainly driven by benchmarks that are disconnected from the specifics of AR.
They mostly evaluate localization and mapping using single still images and either lack temporal changes~\cite{shotton2013scene,advio} or accurate ground truth~(GT)~\cite{sattler2018benchmarking,kendall2015,taira2018inloc}, are restricted to small scenes~\cite{Balntas2017HPatches,shotton2013scene,kendall2015,wald2020,schops2017multi} or landmarks~\cite{Jin2020Image,Schonberger2017Comparative} with perfect coverage and limited viewpoint variability, or disregard temporal tracking data or additional visual, inertial, or radio sensors~\cite{sattler2012aachen,sattler2018benchmarking,taira2018inloc,lee2021naver,nclt,sun2017dataset}.

Our first contribution is to introduce {\bf a large-scale dataset captured using AR devices in diverse environments}, notably a historical building, a multi-story office building, and part of a city center.
The initial data release contains both indoor and outdoor images with illumination and semantic changes as well as dynamic objects.
Specifically, we collected multi-sensor data streams (images, depth, tracking, IMU, BT, WiFi) totalling more than 100 hours using head-mounted HoloLens 2 and hand-held iPhone / iPad devices covering 45'000 square meters over the span of one year (\cref{fig:teaser}).

Second, we develop {\bf a GT pipeline to automatically and accurately register AR trajectories} against large-scale 3D laser scans.
Our pipeline does not require any manual labelling or setup of custom infrastructure (\eg, fiducial markers).
Furthermore, the system robustly handles crowd-sourced data from heterogeneous devices captured over longer periods of time and can be easily extended to support future devices.

Finally, we present {\bf a rigorous evaluation of localization and mapping in the context of AR} and provide {\bf novel insights for future research}.
Notably, we show that the performance of state-of-the-art methods can be drastically improved by considering additional data streams generally available in AR devices, such as radio signals or sequence odometry.
Thus, future algorithms in the field of AR localization and mapping should always consider these sensors in their evaluation to show real-world impact.

The LaMAR dataset, benchmark, GT pipeline, and the implementations of baselines integrating additional sensory data are all publicly available at~\href{https://lamar.ethz.ch/}{\texttt{lamar.ethz.ch}}.
We hope that this will spark future research addressing the challenges of AR.

\section{Related work}

\begin{table*}[t]
\centering
\resizebox{\textwidth}{!}{%
\scriptsize{\setlength\tabcolsep{2pt}%
\newcommand{\good}{\cellcolor{green!25}}%
\newcommand{\bad}{\cellcolor{red!25}}%
\newcommand{\ok}{\cellcolor{gray!25}}%
\newcommand{\zeroStar}{\faIcon[regular]{star}\faIcon[regular]{star}\faIcon[regular]{star}}%
\newcommand{\halfStar}{\faIcon[solid]{star-half-alt}\faIcon[regular]{star}\faIcon[regular]{star}}%
\newcommand{\oneStar}{\faIcon{star}\faIcon[regular]{star}\faIcon[regular]{star}}%
\newcommand{\onehalfStar}{\faIcon{star}\faIcon[solid]{star-half-alt}\faIcon[regular]{star}}%
\newcommand{\twoStar}{\faIcon{star}\faIcon{star}\faIcon[regular]{star}}%
\newcommand{\twohalfStar}{\faIcon{star}\faIcon{star}\faIcon[solid]{star-half-alt}}%
\newcommand{\threeStar}{\faIcon{star}\faIcon{star}\faIcon{star}}%
\begin{tabular}{lccccccccc}
    \toprule
    dataset & out/indoor & changes & scale & density & camera motion & imaging devices & additional sensors & ground truth & accuracy\\
    \midrule
    Aachen~\cite{sattler2012aachen,sattler2018benchmarking} & \cmark\ \xmark & \iconNight\iconStruct & \twohalfStar & \twoStar& \bad still images & DSLR & \xmark & SfM & \bad$>$dm\\
    Phototourism~\cite{Jin2020Image} & \cmark\ \xmark & \iconDynamic\iconStruct & \halfStar & \threeStar & \bad still images & DSLR, phone & \xmark & SfM & \bad$\sim$m\\
    San Francisco~\cite{sanfrancisco} & \cmark\ \xmark & \iconDynamic\iconStruct & \threeStar & \oneStar & \bad still images & DSLR, phone & GNSS & SfM+GNSS & \bad$\sim$m\\
    Cambridge~\cite{kendall2015} & \cmark\ \xmark & \iconDynamic\iconWeather & \halfStar & \twoStar & \ok handheld & mobile & \xmark & SfM & \bad$>$dm\\
    7Scenes~\cite{shotton2013scene} & \xmark\ \cmark & \xmark & \halfStar & \threeStar & \ok handheld & mobile & depth & RGB-D & \good$\sim$cm\\
    RIO10~\cite{wald2020} & \xmark\ \cmark & \iconChair & \halfStar & \threeStar & \ok handheld & Tango tablet & depth & VIO & \bad$>$dm\\
    InLoc~\cite{taira2018inloc} & \xmark\ \cmark & \iconChair & \onehalfStar & \halfStar & \bad still images & panoramas, phone & lidar & manual+lidar & \bad$>$dm\\
    Baidu mall~\cite{sun2017dataset} & \xmark\ \cmark & \iconDynamic & \onehalfStar & \twoStar & \bad still images & DSLR, phone & lidar & manual+lidar & \ok$\sim$dm\\
    Naver Labs~\cite{lee2021naver} & \xmark\ \cmark & \iconDynamic\iconChair & \twoStar & \twoStar & \bad robot-mounted & fisheye, phone & lidar & lidar+SfM & \ok$\sim$dm\\
    NCLT~\cite{nclt} & \cmark\ \cmark & \iconWeather\iconChair & \twoStar & \twoStar & \bad robot-mounted & wide-angle & lidar, IMU, GNSS & lidar+VIO & \ok$\sim$dm\\
    ADVIO~\cite{advio} & \cmark\ \cmark & \iconDynamic & \twoStar & \halfStar & \ok handheld & phone, Tango & IMU, depth, GNSS & manual+VIO & \bad$\sim$m\\
    ETH3D~\cite{schops2017multi} & \cmark\ \cmark & \xmark & \halfStar & \twoStar & \ok handheld & DSLR, wide-angle & lidar & manual+lidar & \good$\sim$mm\\
    \midrule
    \b{LaMAR (ours)} & \cmark\ \cmark
    & \makecell{\iconDynamic\iconWeather\iconNight\\\iconChair\iconStruct}
    & \makecell{\twohalfStar\\3 locations \\ 45'000 m${}^2$}
    & \makecell{\threeStar\\100 hours\\40 km}
    & \good\makecell{handheld\\head-mounted} 
    & \makecell{phone, headset \\ backpack, trolley} 
    & \makecell{lidar, IMU, \faIcon{wifi}\ \faIcon{bluetooth}\\ depth, infrared} 
    & \makecell{lidar+SfM+VIO\\automated} & \good$\sim$cm\\
    \bottomrule
\end{tabular}
}%
}%
\vspace{1mm}
\caption{\textbf{Overview of existing datasets.}
No dataset, besides ours, exhibits at the same time short-term appearance and structural changes due to moving people~\iconDynamic, weather~\iconWeather, or day-night cycles~\iconNight, but also long-term changes due to displaced furniture~\iconChair\ or construction work~\iconStruct. 
}%
\label{tab:datasets}
\end{table*}

\PAR{Image-based localization}
is classically tackled by estimating a camera pose from correspondences established between sparse local features~\cite{lowe2004distinctive,bay2008speeded,Rublee2011ORB,mikolajczyk2004ijcv} and a 3D Structure-from-Motion (SfM)~\cite{Schoenberger2016Structure} map of the scene~\cite{fischler1981random,li2012worldwide,sattler2012improving}.
This pipeline scales to large scenes using image retrieval~\cite{arandjelovic2012three,vlad,apgem,asmk,cao2020unifying,rau2020imageboxoverlap,densevlad}.
Recently, many of these steps or even the end-to-end pipeline have been successfully learned with neural networks~\cite{detone2018superpoint,sarlin2020superglue,Dusmanu2019CVPR,schoenberger2018semantic,arandjelovic2016netvlad,NIPS2017_831caa1b,tian2019sosnet,sarlin2019,yi2016lift,Hyeon2021,sarlin21pixloc,lindenberger2021pixsfm}.
Other approaches regress absolute camera pose~\cite{kendall2015,kendall2017geometric,ng2021reassessing} or scene coordinates~\cite{shotton2013scene,valentin2015cvpr,meng2017backtracking,massiceti2017random,angle_scr,brachmann2019esac,Wang2021,brachmann2021dsacstar}.
However, all these approaches typically fail whenever there is lack of context (\eg, limited field-of-view) or the map has repetitive elements.
Leveraging the sequential ordering of video frames~\cite{seqslam,johns2013feature} or modelling the problem as a generalized camera~\cite{pless2003using,hee2016minimal,sattler2018benchmarking,speciale2019a} can improve results. 

\PAR{Radio-based localization:}
Radio signals, such as WiFi and Bluetooth, are spatially bounded (logarithmic decay)~\cite{radar,khalajmehrabadi2016modern,radio_fingerprint}, thus can distinguish similarly looking (spatially distant) locations.
Their unique identifiers can be uniquely hashed which makes them computationally attractive (compared with high-dimensional image descriptors).
Several methods use the signal strength, angle, direction, or time of arrival~\cite{radio_aoa,radio_toa,radio_tdoa} but the most popular is model-free map-based fingerprinting~\cite{khalajmehrabadi2016modern,radio_fingerprint,fault_tolerant}, as it only requires to collect unique identifiers of nearby radio sources and received signal strength.
GNSS provides absolute 3-DoF positioning but is not applicable indoors and has insufficient accuracy for AR scenarios, especially in urban environments due to multi-pathing, \etc

\PAR{Datasets and ground-truth:}
Many of the existing benchmarks (\cf~\cref{tab:datasets}) are captured in small-scale environments \cite{shotton2013scene,wald2020,dai2017scannet,hodan2018bop}, do not contain sequential data~\cite{sattler2012aachen,Jin2020Image,sanfrancisco,taira2018inloc,sun2017dataset,schops2017multi,Balntas2017HPatches,Schonberger2017Comparative}, lack characteristic hand-held/head-mounted motion patterns ~\cite{sattler2018benchmarking,Badino2011,RobotCarDatasetIJRR,wenzel2020fourseasons}, or their GT is not accurate enough for AR~\cite{advio,kendall2015}.
None of these datasets contain WiFi or Bluetooth data (\cref{tab:datasets}). 
The closest to our work are Naver Labs~\cite{lee2021naver}, NCLT~\cite{nclt} and ETH3D~\cite{schops2017multi}. 
Both, Naver Labs~\cite{lee2021naver} and NCLT~\cite{nclt} are less accurate than ours and do not contain AR specific trajectories or radio data. 
The Naver Labs dataset~\cite{lee2021naver} also does not contain any outdoor data. 
ETH3D~\cite{schops2017multi} is highly accurate, however, it is only small-scale, does not contain significant changes, or any radio data.

To establish ground-truth, many datasets rely on off-the-shelf SfM algorithms~\cite{Schoenberger2016Structure} for unordered image collections~\cite{sattler2012aachen,Jin2020Image,kendall2015,wald2020,advio,sun2017dataset,taira2018inloc,Jin2020Image}.
Pure SfM-based GT generation has limited accuracy~\cite{brachmann2021limits} and completeness, which biases the evaluations to scenarios in which visual localization already works well.
Other approaches rely on RGB(-D) tracking~\cite{wald2020,shotton2013scene}, which usually drifts in larger scenes and cannot produce GT in crowd-sourced, multi-device scenarios. 
Specialized capture rigs of an AR device with a more accurate sensor (lidar)~\cite{lee2021naver,nclt} prevent capturing of realistic AR motion patterns.
Furthermore, scalability is limited for these approaches, especially if they rely on manual selection of reference images~\cite{sun2017dataset}, laborious labelling of correspondences~\cite{sattler2012aachen,taira2018inloc}, or placement of fiducial markers~\cite{hodan2018bop}.
For example, the accuracy of ETH3D~\cite{schops2017multi} is achieved by using single stationary lidar scan, manual cleaning, and aligning very few images captured by tripod-mounted DSLR cameras.
Images thus obtained are not representative for AR devices and the process cannot scale or take advantage of crowd-sourced data.
In contrast, our fully automatic approach does not require any manual labelling or special capture setups, thus enables light-weight and repeated scanning of large locations.

\begin{table*}[t]
\centering
\scriptsize{\setlength\tabcolsep{1.7pt}%
\begin{tabular}{ccccccccccc}
    \toprule
    \multirow{2}{*}{device} & \multirow{2}{*}{\makecell{motion\\type}} & \multicolumn{5}{c}{cameras} & \multirow{2}{*}{radios} & \multirow{2}{*}{other data} & \multirow{2}{*}{poses}\\
    \cmidrule(lr){3-7}
    & & \# & FOV & frequency & resolution & specs\\
    \midrule
    M6 & trolley & 6 & 113\degree & 1-3m & 1080p & RGB, sync & \faIcon{wifi}\faIcon{bluetooth} & lidar points+mesh & lidar SLAM\\
    VLX & backpack & 4 & 90\degree & 1-3m & 1080p & RGB, sync & \faIcon{bluetooth} & lidar points+mesh & lidar SLAM\\
    HoloLens2 & head-mounted & 4 & 83\degree & 30Hz & VGA & gray, GS & \faIcon{wifi}\faIcon{bluetooth} & ToF depth/IR 1Hz, IMU & head-tracking\\
    iPad/iPhone & hand-held & 1 & 64\degree & 10Hz & 1080p & RGB, RS, AF & \faIcon{bluetooth}${}^*$ & lidar depth 10Hz, IMU & ARKit\\
    \bottomrule
\end{tabular}}
\vspace{1mm}%
\caption{\textbf{Sensor specifications.}
Our dataset has visible light images (global shutter GS, rolling shutter RS, auto-focus AF), depth data (ToF, lidar), radio signals (${}^*$, if partial), dense lidar point clouds, and poses with intrinsics from on-device tracking. %
}%
\label{tab:sensors}
\end{table*}

\section{Dataset}
\label{sec:data}

We first give an overview of the setup and content of our dataset.

\PAR{Locations:}
The initial release of the dataset contains 3 large locations representative of AR use cases:
1) HGE (18'000 m$^2$) is the ground floor of a historical university building composed of multiple large halls and large esplanades on both sides.
2) CAB (12'000 m$^2$) is a multi-floor office building composed of multiple small and large offices, a kitchen, storage rooms, and 2 courtyards.
3) LIN (15'000 m$^2$) is a few blocks of an old town with shops, restaurants, and narrow passages. 
HGE and CAB contain both indoor and outdoor sections with many symmetric structures.
Each location underwent structural changes over the span of a year, \eg, the front of HGE turned into a construction site and the indoor furniture was rearranged.
See \cref{fig:locations} and \cref{sec:visuals} for visualizations.

\PAR{Data collection:}
We collected data using Microsoft HoloLens 2 and Apple iPad Pro devices with custom raw sensor recording applications.
10 participants were each given one device and asked to walk through a common designated area.
They were only given the instructions to freely walk through the environment to visit, inspect, and find their way around.
This yielded diverse camera heights and motion patterns.
Their trajectories were not planned or restricted in any way.
Participants visited each location, both during the day and at night, at different points in time over the course of up to 1 year.
In total, each location is covered by more than 100 sessions of 5 minutes.
We did not need to prepare the capturing site in any way before recording.
This enables easy barrier-free crowd-sourced data collections. %
Each location was also captured two to three times by NavVis M6 trolley or VLX backpack mapping platforms, which generate textured dense 3D models of the environment using laser scanners and panoramic cameras.

\PAR{Privacy:}
We paid special attention to comply with privacy regulations. 
Since the dataset is recorded in public spaces, our pipeline anonymizes all visible faces and licence plates.

\PAR{Sensors:}
We provide details about the recorded sensors in \cref{tab:sensors}.
The HoloLens has a specialized large field-of-view (FOV) multi-camera tracking rig (low resolution, global shutter) \cite{ungureanu2020hololens}, while the iPad has a single, higher-resolution camera with rolling shutter and more limited FOV.
We also recorded outputs of the real-time AR tracking algorithms available on each device, which includes relative camera poses and sensor calibration.
All images are undistorted.
All sensor data is registered into a common reference frame with accurate absolute GT poses using the pipeline described in the next section.

\begin{figure}[t]
    \centering
    \input{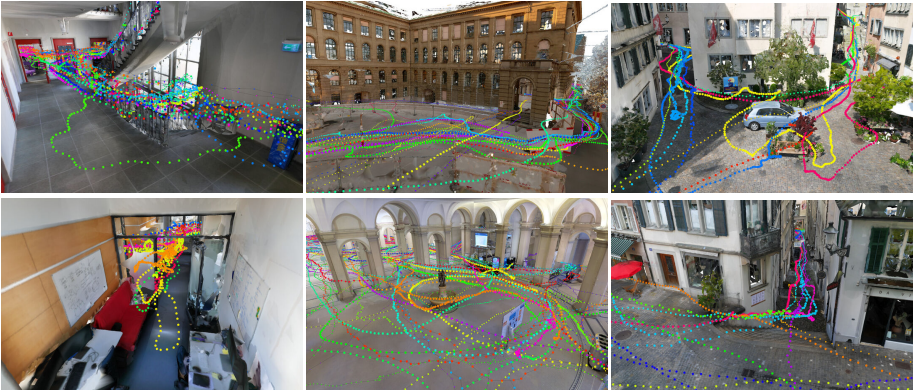}
    \vspace{2mm}
    \caption{%
    \textbf{The locations feature diverse indoor and outdoor spaces.}
    High-quality meshes, obtained from lidar, are registered with numerous AR sequences, each shown here as a different color.
    }
    \label{fig:locations}
\end{figure}

\section{Ground-truth generation}
\label{sec:method}
The GT estimation process takes as input the raw data from the different sensors.
The entire pipeline is fully automated and does not require any manual alignment or input.

\PAR{Overview:}
We start by aligning different sessions of the laser scanner by using the images and the 3D lidar point cloud.
When registered together, they form the GT reference map, which accurately captures the structure and appearance of the scene.
We then register each AR sequence individually to the reference map using local feature matching and relative poses from the on-device tracker.
Finally, all camera poses are refined jointly by optimizing the visual constraints within and across sequences.

\PAR{Notation:}
We denote ${}_i\*T_j \in \text{SE}(3)$ the 6-DoF pose, encompassing rotation and translation, that transforms a point in frame $j$ to another frame $i$.
Our goal is to compute globally-consistent absolute poses ${}_w\*T_i$ for all cameras $i$ of all sequences and scanning sessions into a common reference world frame $w$.

\subsection{Ground-truth reference model}
\label{sec:method:ref}

Each capture session $S \in \mathcal S$ of the NavVis laser-scanning platform is processed by a proprietary inertial-lidar SLAM that estimates, for each image $i$, a pose ${}_0\*T_i^S$ relative to the beginning of the session.
The software filters out noisy lidar measurements, removes dynamic objects, and aggregates the remainder into a globally-consistent colored 3D point cloud with a grid resolution of 1cm.
To recover visibility information, we compute a dense mesh using the Advancing Front algorithm~\cite{cohen2004greedy}.

Our first goal is to align the sessions into a common GT reference frame.
We assume that the scan trajectories are drift-free and only need to register each with a rigid transformation~${}_w\*T_0^S$.
Scan sessions can be captured between extensive periods of time and therefore exhibit large structural and appearance changes.
We use a combination of image and point cloud information to obtain accurate registrations without any manual initialization.
The steps are inspired by the reconstruction pipeline of Choi~\etal~\cite{choi2015robust,Zhou2018}.

\PAR{Pair-wise registration:}
We first estimate a rigid transformation ${}_A\*T_B$ for each pair of scanning sessions $(A, B) \in \mathcal{S}^2$.
For each image $I_i^A$ in $A$, we select the $r$ most similar images $(I^B_j)_{1 \leq j \leq r}$ in $B$ based on global image descriptors~\cite{vlad,arandjelovic2016netvlad,apgem}, which helps the registration scale to large scenes.
We extract sparse local image features and establish 2D-2D correspondences $\{\*p^A_i,\*p^B_j\}$ for each image pair $(i, j)$.
The 2D keypoints $\*p_i \in \mathbb{R}^2$ are lifted to 3D, $\*P_i \in \mathbb{R}^3$, by tracing rays through the dense mesh of the corresponding session.
This yields 3D-3D correspondences $\{\*P^A_i,\*P^B_j\}$, from which we estimate an initial relative pose~\cite{umeyama1991least} using RANSAC~\cite{fischler1981random}.
This pose is refined with the point-to-plane Iterative Closest Point (ICP) algorithm~\cite{rusinkiewicz2001efficient} applied to the pair of lidar point clouds.

We use state-of-the-art local image features that can match across drastic illumination and viewpoint changes~\cite{sarlin2019,detone2018superpoint,revaudr2d2}.
Combined with the strong geometric constraints in the registration, our system is robust to long-term temporal changes and does not require manual initialization.
Using this approach, we have successfully registered building-scale scans captured at more than a year of interval with large structural changes.

\PAR{Global alignment:}
We gather all pairwise constraints and jointly refine all absolute scan poses $\{{}_w\*T_0^S\}$ by optimizing a pose graph~\cite{grisetti2010tutorial}.
The edges are weighted with the covariance matrices of the pair-wise ICP estimates.
The images of all scan sessions are finally combined into a unique reference trajectory $\{{}_w\*T_i^\text{ref}\}$.
The point clouds and meshes are aligned according to the same transformations.
They define the reference representation of the scene, which we use as a basis to obtain GT for the AR sequences.

\PAR{Ground-truth visibility:}
The accurate and dense 3D geometry of the mesh allows us to compute accurate visual overlap between two cameras with known poses and calibration.
Inspired by Rau~\etal~\cite{rau2020imageboxoverlap}, we define the overlap of image $i$ wrt.~a reference image $j$ by the ratio of pixels in $i$ that are visible in $j$:
\begin{equation}
    O(i\rightarrow j) = \frac{\sum_{k\in(W,H)} \mathbbm{1}\left[
        \mathrm{\Pi}_j({}_w\*T_j, \mathrm{\Pi}_i^{-1}({}_w\*T_i, \*p^i_k, z_k)) \in (W, H)
    \right]\:\alpha_k
    }{W\cdot H} \enspace,
\end{equation}
where $\mathrm{\Pi}_i$ projects a 3D point $k$ to camera $i$, $\mathrm{\Pi}_i^{-1}$ conversely backprojects it using its known depth $z_k$ with $(W, H)$ as the image dimensions.
The contribution of each pixel is weighted by the angle $\alpha_k = \cos(\*n_{i,k}, \*n_{j,k})$ between the two rays.
To handle scale changes, it is averaged both ways $i\rightarrow j$ and $j\rightarrow i$.
This score is efficiently computed by tracing rays through the mesh and checking for occlusion for robustness.

This score $O\in[0,1]$ favors images that observe the same scene from similar viewpoints.
Unlike sparse co-visibility in an SfM model~\cite{radenovic2018fine}, our formulation is independent of the amount of texture and the density of the feature detections.
This score correlates with matchability -- we thus use it as GT when evaluating retrieval and to determine an upper bound on the theoretically achievable performance of our benchmark.

\subsection{Sequence-to-scan alignment}
\label{sec:method:seq}
We now aim to register each AR sequence individually into the dense GT reference model (see \cref{fig:seq-to-scan}).
Given a sequence of $n$ frames, we introduce a simple algorithm that estimates the per-frame absolute pose $\{{}_w\*T_i\}_{1 \leq i \leq n}$.
A frame refers to an image taken at a given time or, when the device is composed of a camera rig with known calibration (\eg, HoloLens), to a collection of simultaneously captured images.

\begin{figure}[t]
    \centering
    \includegraphics[width=\textwidth]{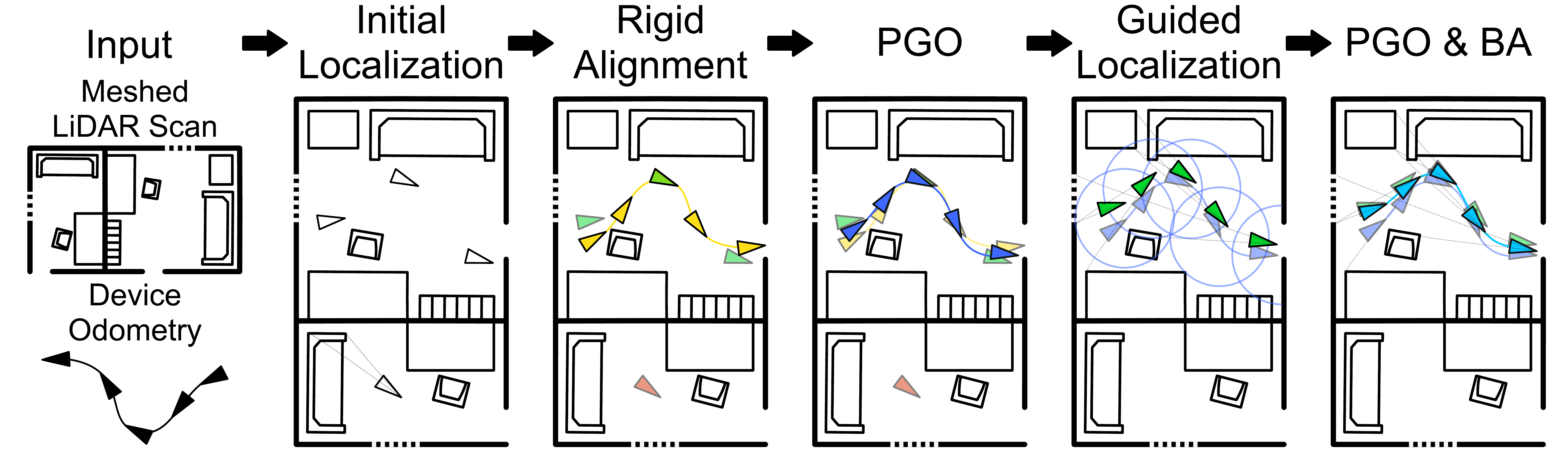}
    \caption{{\bf Sequence-to-scan alignment.}
    We first estimate the absolute pose of each sequence frame using image retrieval and matching.
    This initial localization prior is used to obtain a single rigid alignment between the input trajectory and the reference 3D model via voting.
    The alignment is then relaxed by optimizing the individual frame poses in a pose graph based on both relative and absolute pose constraints.
    We bootstrap this initialization by mining relevant image pairs and re-localizing the queries.
    Given these improved absolute priors, we optimize the pose graph again and finally include reprojection errors of the visual correspondences, yielding a refined trajectory.
    }
    \label{fig:seq-to-scan}
\end{figure}

\PAR{Inputs:}
We assume given trajectories $\{{}_0\*T^\text{track}_{i}\}$ estimated by a visual-inertial tracker -- we use ARKit for iPhone/iPad and the on-device tracker for HoloLens.
The tracker also outputs per-frame camera intrinsics $\{\*C_i\}$, which account for auto-focus or calibration changes and are for now kept fixed.

\PAR{Initial localization:}
For each frame of a sequence $\{I^{\text{query}}_{i}\}$, we retrieve a fixed number $r$ of relevant reference images $(I^\text{ref}_j)_{1 \leq j \leq r}$ using global image descriptors.
We match sparse local features~\cite{lowe2004distinctive,detone2018superpoint,revaudr2d2} extracted in the query frame to each retrieved image $I^\text{ref}_j$ obtaining a set of 2D-2D correspondences $\{\*p_{i,k}^\text{q},\*p_{j,k}^\text{ref}\}_k$.
The 2D reference keypoints are lifted to 3D by tracing rays through the mesh of the reference model, yielding a set of 2D-3D correspondences $\mathcal{M}_{i, j} : = \{\*p_{i,k}^\text{q},\*P_{j,k}^\text{ref}\}_k$.
We combine all matches per query frame $\mathcal{M}_{i} = \cup_{j=1}^{r} \mathcal{M}_{i, j}$ and estimate an initial absolute pose ${}_w\*T^\text{loc}_i$ using the (generalized) P3P algorithm~\cite{hee2016minimal} within a LO-RANSAC scheme~\cite{chum2003locally} followed by a non-linear refinement~\cite{Schoenberger2016Structure}.
Because of challenging appearance conditions, structural changes, or lack of texture, some frames cannot be localized in this stage.
We discard all poses that are supported by a low number of inlier correspondences.

\PAR{Rigid alignment:}
We next recover a coarse initial pose $\{{}_w\*T^\text{init}_i\}$ for all frames, including those that could not be localized.
Using the tracking, which is for now assumed drift-free, we find the rigid alignment ${}_w\*T^\text{init}_0$ that maximizes the consensus among localization poses.
This voting scheme is fast and effectively rejects poses that are incorrect, yet confident, due to visual aliasing and symmetries.
Each estimate is a candidate transformation ${}_w\*T^i_0 = {}_w\*T^\text{loc}_i \left({}_0\*T^\text{track}_{i}\right)^{-1}$, for which other frames can vote, if they are consistent within a threshold $\tau_\text{rigid}$.
We select the candidate with the highest count of inliers:
\begin{equation}
    {}_w\*T^\text{init}_0 = \argmax_{\*T \in \{{}_w\*T^i_0\}_{1 \leq i \leq n}} \sum_{1 \leq j \leq n} 
    \mathbbm{1}\left[\text{dist} \left( {}_w\*T^\text{loc}_j, \*T\cdot{}_0\*T^\text{track}_{j} \right) < \tau_\text{rigid}\right] \enspace,
\end{equation}
where $\mathbbm{1}\left[\cdot\right]$ is the indicator function and $\text{dist} \left( \cdot, \cdot \right)$ returns the magnitude, in terms of translation and rotation, of the difference between two absolute poses.
We then recover the per-frame initial poses as $\{{}_w\*T^\text{init}_i := {}_w\*T^\text{init}_0 \cdot {}_0\*T^\text{track}_{i}\}_{1 \leq i \leq n}$.

\PAR{Pose graph optimization:}
We refine the initial absolute poses by maximizing the consistency of tracking and localization cues within a pose graph.
The refined poses $\{{}_w\*T^\text{PGO}_i\}$ minimize the energy function
\begin{equation}
    E(\{{}_w\*T_i\}) =
    \sum_{i = 1}^{n - 1} \mathcal{C}_\text{PGO}\left({}_w\*T_{i+1}^{-1}\:{}_w\*T_i,\ {}_{i+1}\*T^\text{track}_{i} \right)
    + \sum_{i = 1}^n \mathcal{C}_\text{PGO}\left({}_w\*T_i,\ {}_w\*T^\text{loc}_i\right) \enspace,
\end{equation}
where $\mathcal{C}_\text{PGO}\left(\*T_1, \*T_2\right) := \left\Vert\text{Log}\left(\*T_1\:\*T_2^{-1}\right)\right\Vert^2_{\Sigma,\gamma}$ is the distance between two absolute or relative poses, weighted by covariance matrix $\Sigma \in \mathbb{R}^{6\times6}$ and loss function $\gamma$. 
Here, $\text{Log}$ maps from the Lie group $\text{SE}(3)$ to the corresponding algebra $\mathfrak{se}(3)$.

We robustify the absolute term with the Geman-McClure loss function and anneal its scale via a Graduated Non-Convexity scheme~\cite{yang2020graduated}.
This ensures convergence in case of poor initialization, \eg, when the tracking exhibits significant drift, while remaining robust to incorrect localization estimates.
The covariance of the absolute term is propagated from the preceding non-linear refinement performed during localization.
The covariance of the relative term is recovered from the odometry pipeline, or, if not available, approximated as a factor of the motion magnitude.

This step can fill the gaps from the localization stage using the tracking information and conversely correct for tracker drift using localization cues.
In rare cases, the resulting poses might still be inaccurate when both the tracking drifts and the localization fails.

\PAR{Guided localization via visual overlap:}
To further increase the pose accuracy, we leverage the current pose estimates $\{{}_w\*T^\text{PGO}_i\}$ to mine for additional localization cues.
Instead of relying on global visual descriptors, which are easily affected by aliasing, we select reference images with a high overlap using the score defined in \cref{sec:method:ref}.
For each sequence frame $i$, we select $r$ reference images with the largest overlap and again match local features and estimate an absolute pose.
These new localization priors improve the pose estimates in a second optimization of the pose graph.

\PAR{Bundle adjustment:}
For each frame $i$, we recover the set of 2D-3D correspondences $\mathcal{M}_i$ used by the guided re-localization.
We now refine the poses $\{{}_w\*T^\text{BA}_i\}$ by jointly minimizing a bundle adjustment problem with relative pose graph costs:
\begin{align}
\begin{split}
    E(\{{}_w\*T_i\}) =
    & \sum_{i = 1}^{n - 1} \mathcal{C}_\text{PGO}\left({}_w\*T_{i+1}^{-1}\:{}_w\*T_i,\ {}_{i+1}\*T^\text{track}_{i} \right) \\
    +& \sum_{i=1}^{n} \sum_{\mathcal{M}_{i,j}\in \mathcal{M}_{i}}\sum_{(\*p^\text{ref}_k, \*P^\text{q}_k) \in \mathcal M_{i,j}}
        \left\Vert\mathrm{\Pi} ({}_w\*T_i, \*P^\text{ref}_{j,k}) - \*p^\text{q}_{i,k}\right\Vert^2_{\sigma^2} \enspace,
\label{eq:pgo-ba}
\end{split}
\end{align}
where the second term evaluates the reprojection error of a 3D point $\*P^\text{ref}_{j,k}$ for observation $k$ to frame $i$.
The covariance is the noise $\sigma^2$ of the keypoint detection algorithm. %
We pre-filter correspondences that are behind the camera or have an initial reprojection error greater than $\sigma\,\tau_\text{reproj}$.
As the 3D points are sampled from the lidar, we also optimize them with a prior noise corresponding to the lidar specifications.
We use the Ceres~\cite{ceres-solver} solver.

\subsection{Joint global refinement}
\label{sec:refinement}
Once all sequences are individually aligned, we refine them jointly by leveraging sequence-to-sequence visual observations. 
This is helpful when sequences observe parts of the scene not mapped by the LiDAR. 
We first triangulate a sparse 3D model from scan images, aided by the mesh.
We then triangulate additional observations, and finally jointly optimize the whole problem.

\PAR{Reference triangulation:}
We estimate image correspondences of the reference scan using pairs selected according to the visual overlap defined in \cref{sec:method:seq}.
Since the image poses are deemed accurate and fixed, we filter the correspondences using the known epipolar geometry.
We first consider feature tracks consistent with the reference surface mesh before triangulating more noisy observations within LO-RANSAC using COLMAP~\cite{Schoenberger2016Structure}.
The remaining feature detections, which could not be reliably matched or triangulated, are lifted to 3D by tracing through the mesh.
This results in an accurate, sparse SfM model with tracks across reference images.

\PAR{Sequence optimization:}
We then add each sequence to the sparse model.
We first establish correspondences between images of the same and of different sequences.
The image pairs are again selected by highest visual overlap computed using the aligned poses $\{{}_w\*T^\text{BA}_i\}$.
The resulting tracks are sequentially triangulated, merged, and added to the sparse model.
Finally, all 3D points and poses are jointly optimized by minimizing the joint pose-graph and bundle adjustment (\cref{eq:pgo-ba}).
As in COLMAP~\cite{Schoenberger2016Structure}, we alternate optimization and track merging.
To scale to large scenes, we subsample keyframes from the full frame-rate captures and only introduce absolute pose and reprojection constraints for keyframes while maintaining all relative pose constraints from tracking.

\begin{figure}[t]
    \centering
    \begin{minipage}{0.52\textwidth}
        \includegraphics[width=\linewidth]{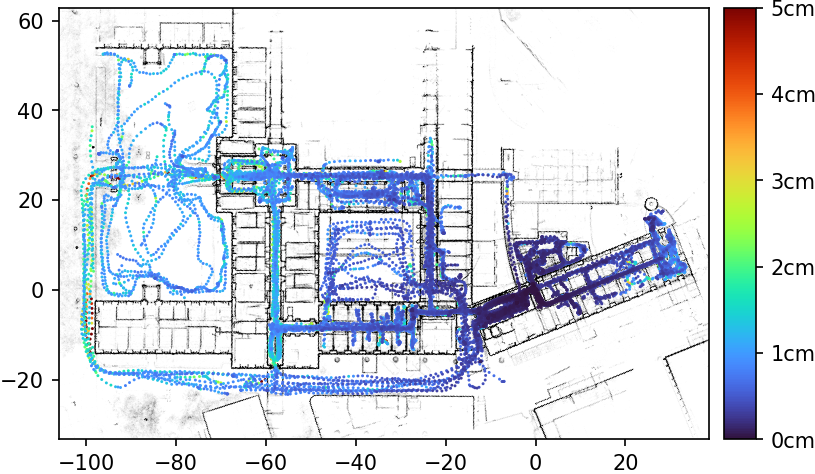}
    \end{minipage}%
    \hspace{0.01\textwidth}%
    \begin{minipage}{0.46\textwidth}
        \centering
        \includegraphics[width=.49\textwidth]{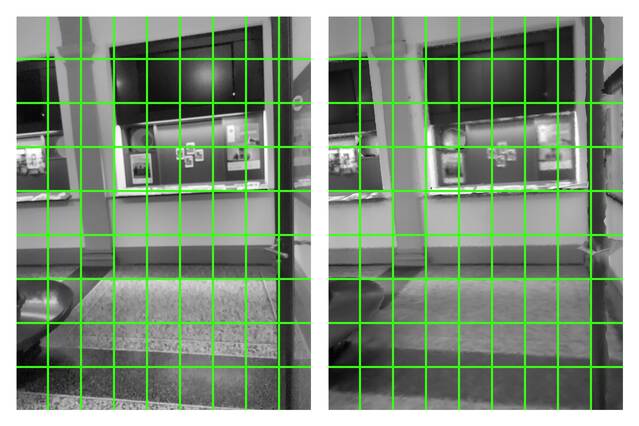}%
        \hspace{0.01\textwidth}%
        \includegraphics[width=.49\textwidth]{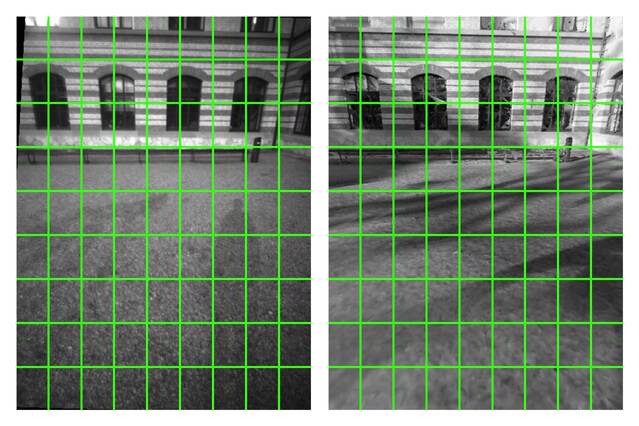}
        
        \includegraphics[width=.49\linewidth]{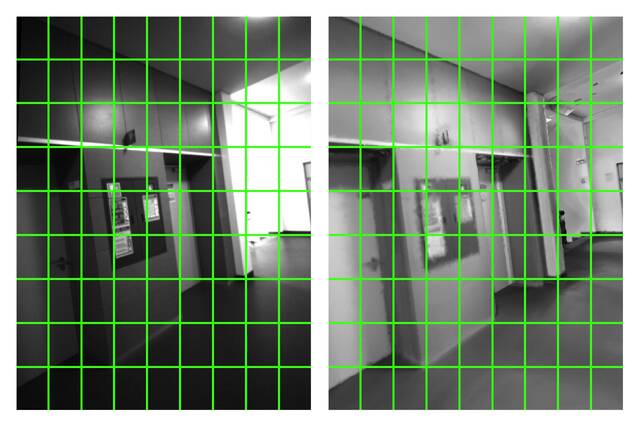}%
        \hspace{0.01\textwidth}%
        \includegraphics[width=.49\linewidth]{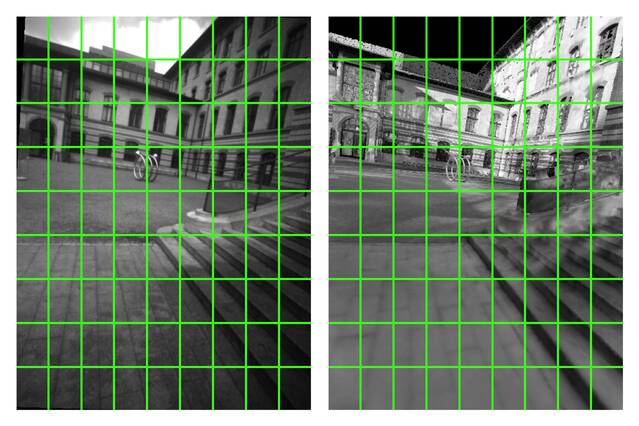}
    \end{minipage}
    \vspace{2mm}
    \caption{%
    \textbf{Uncertainty of the GT poses for the CAB scene.}
    Left: The overhead map shows that the translation uncertainties are larger in long corridors and outdoor spaces.
    Right: Pairs of captured images (left) and renderings of the mesh at the estimated camera poses (right). They are pixel-aligned, which confirms that the poses are sufficiently accurate for our evaluation.
    }%
    \label{fig:uncertainties}%
\end{figure}

\subsection{Ground-truth validation}

\PAR{Potential limits:}
Brachmann~\etal~\cite{brachmann2021limits} observe that algorithms generating pseudo-GT poses by minimizing either 2D or 3D cost functions alone can yield noticeably different results.
We argue that there exists a single underlying, true GT.
Reaching it requires fusing large amounts of redundant data with sufficient sensors of sufficiently low noise.
Our GT poses optimize complementary constraints from visual and inertial measurements, guided by an accurate lidar-based 3D structure.
Careful design and propagation of uncertainties reduces the bias towards one of the sensors.
All sensors are factory- and self-calibrated during each recording by the respective commercial, production-grade SLAM algorithms.
We do not claim that our GT is perfect but analyzing the optimization uncertainties sheds light on its degree of accuracy.

\PAR{Pose uncertainty:}
We estimate the uncertainties of the GT poses by inverting the Hessian of the refinement.
To obtain calibrated covariances, we scale them by the empirical keypoint detection noise, estimated as $\sigma{=}1.33$ pixels for the CAB scene.
The maximum noise in translation is the size of the major axis of the uncertainty ellipsoids, which is the largest eivenvalue $\sigma_t^2$ of the covariance matrices.
\Cref{fig:uncertainties} shows its distribution for the CAB scene.
We retain images whose poses are correct within $10$cm with a confidence of $99.7$\%.
For normally distributed errors, this corresponds to a maximum uncertainty $\sigma_t{=}3.33\text{cm}$ and discards $0.8$\% of all frames.
For visual inspection, we render images at the estimated GT camera poses using the colored mesh.
They appear pixel-aligned with the original images, supporting that the poses are accurate.
We provide additional visualizations in \cref{sec:uncertainties}.

\subsection{Selection of mapping and query sequences}
We divide the set of sequences into two disjoint groups for mapping and localization.
Mapping sequences are selected such that they have a minimal overlap between each other yet cover the area visited by all remaining sequences.
This simulates a scenario of minimal coverage and maximizes the number of localization query sequences.
We cast this as a combinatorial optimization problem solved with a depth-first search.
We provide more details in \cref{sec:map-query-split}.
\section{Evaluation}

We evaluate state-of-the-art approaches in both single-frame and sequence settings and summarize our results in \cref{fig:baselines}.
We build maps using both types of AR devices and evaluate the localization accuracy for 1000 randomly-selected queries of each device for each location.
All results are averaged across all locations.
\Cref{sec:supp:distribution} provides more details about the distribution of the evaluation data.

\PAR{Single-frame:}
We first consider in \cref{sec:eval:single-frame} the classical academic setup of single-frame queries (single image for phones and single rig for HoloLens 2) without additional sensor.
We then look at how radio signals can be beneficial.
We also analyze the impact of various settings: FOV, type of mapping images, and mapping algorithm.

\PAR{Sequence:}
Second, by leveraging the real-time AR tracking poses, we consider the problem of sequence localization in \cref{sec:eval:chunk}.
This corresponds to a real-world AR application retrieving the content attached to a target map using the real-time sensor stream from the device.
In this context, we care not only about accuracy and recall but also about the time required to localize accurately, which we call the \emph{time-to-recall}.

\begin{figure}[t]
\centering
\includegraphics[width=0.70\textwidth]{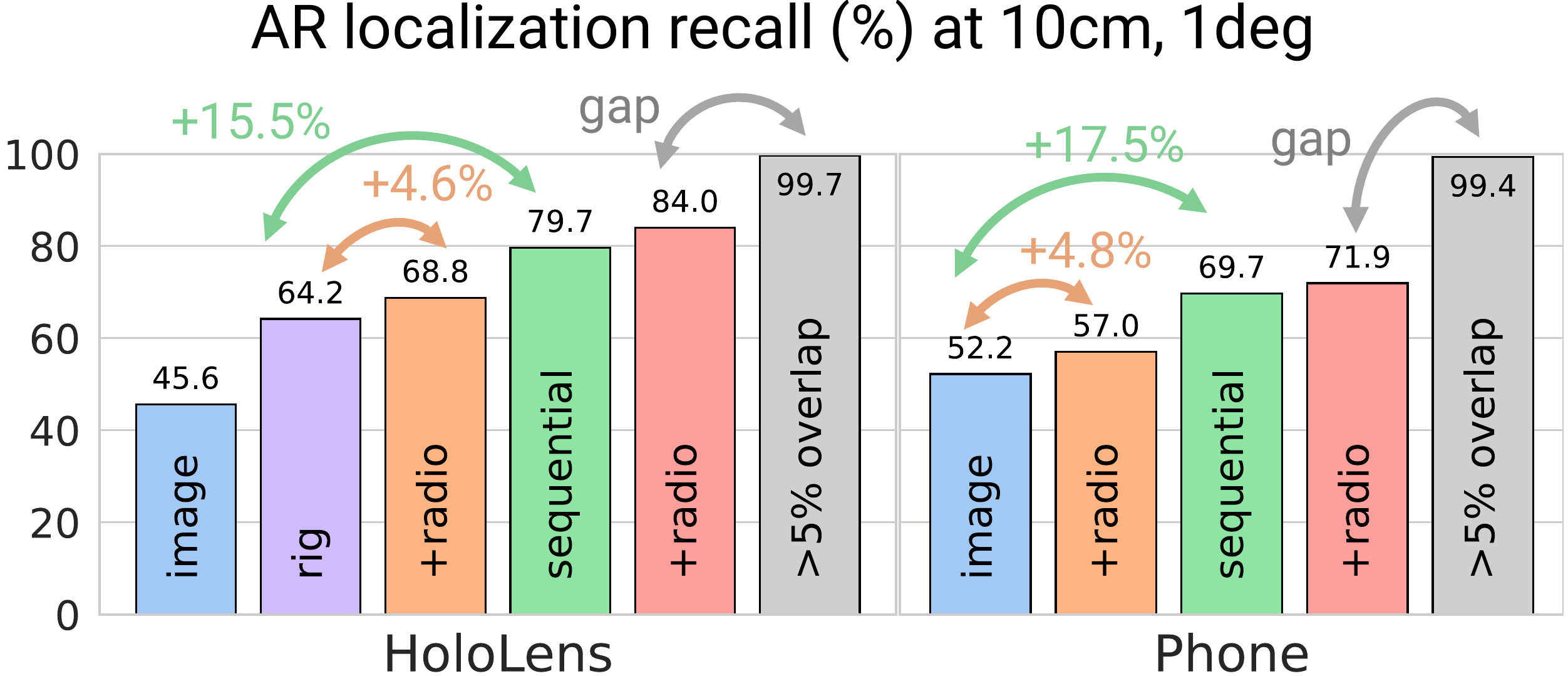}
\caption{\textbf{Main results.} We show results for Fusion image retrieval with SuperPoint local features and SuperGlue matcher on both HoloLens 2 and phone queries.
We consider several tracks: single-image / single-rig localization with / without radios and similarly for sequence (10 seconds) localization.
In addition, we report the percentage of queries with at least 5\% ground-truth overlap with respect to the best mapping image.
}%
\label{fig:baselines}
\end{figure}

\subsection{Single-frame localization}\label{sec:eval:single-frame}

We first evaluate several algorithms representative of the state of the art in the classical single-frame academic setup.
We consider the hierarchical localization framework with different approaches for image retrieval and matching.
Each of them first builds a sparse SfM map from reference images.
For each query frame, we then retrieve relevant reference images, match their local features, lift the reference keypoints to 3D using the sparse map, and finally estimate a pose with PnP+RANSAC.
We report the recall of the final pose at two thresholds~\cite{sattler2018benchmarking}:
1) a fine threshold at ($1^\circ, 10$cm), which we see as the minimum accuracy required for a good AR user experience in most settings.
2) a coarse threshold at ($5^\circ, 1$m) to show the room for improvement for current approaches.

We evaluate global descriptors computed by NetVLAD~\cite{arandjelovic2016netvlad} and by a fusion~\cite{humenberger2020robust} of NetVLAD and APGeM~\cite{apgem}, which are representative of the field~\cite{pion2020benchmarking}.
We retrieve the 10 most similar images.
For matching, we evaluate handcrafted SIFT~\cite{lowe2004distinctive}, SOSNet~\cite{tian2019sosnet} as a learned patch descriptor extracted from DoG~\cite{lowe2004distinctive} keypoints, and a robust deep-learning based joint detector and descriptor R2D2~\cite{revaudr2d2}.
Those are matched by exact mutual nearest neighbor search.
We also evaluate SuperGlue~\cite{sarlin2020superglue} -- a learned matcher based on SuperPoint~\cite{detone2018superpoint} features.
To build the map, we retrieve neighboring images filtered by frustum intersection from reference poses, match these pairs, and triangulate a sparse SfM model using COLMAP~\cite{Schoenberger2016Structure}.

We report the results in \cref{tab:baselines+radio} (left).
Even the best methods have a large gap to perfect scores and much room for improvement.
In the remaining ablation, we solely rely on SuperPoint+SuperGlue~\cite{detone2018superpoint,sarlin2020superglue} for matching as it clearly performs the best.

\begin{table}[t]
\centering
\begin{minipage}{0.47\linewidth}
\scriptsize{\begin{tabular}{cccc}
    \toprule
    \multicolumn{2}{c}{Hierarchical localization} & \multicolumn{2}{c}{Query device} \\
    \cmidrule(lr){1-2} \cmidrule(lr){3-4}
    Retrieval & Matching & HL2 & Phone \\
    \midrule
    \multirow{4}{*}{NetVLAD} & SIFT & 48.3 / 63.7 & 38.0 / 54.8\\
    & DoG+SOSNet & 52.3 / 67.3 & 37.9 / 55.4\\
    & R2D2 & 48.2 / 63.9 & 42.1 / 58.4\\
    & SP+SG & 59.9 / 73.0 & 50.1 / 63.3\\
    \cmidrule{1-4}
    \multirow{4}{*}{Fusion} & SIFT & 51.2 / 67.9 & 38.5 / 56.9\\
    & DoG+SOSNet & 55.2 / 71.2 & 39.3 / 57.4\\
    & R2D2 & 52.0 / 68.4 & 43.5 / 60.2\\
    & SP+SG & 64.2 / 77.4 & 52.2 / 65.8\\
    \bottomrule
\end{tabular}}
\end{minipage}
\begin{minipage}{0.52\linewidth}
\centering
\includegraphics[width=\textwidth]{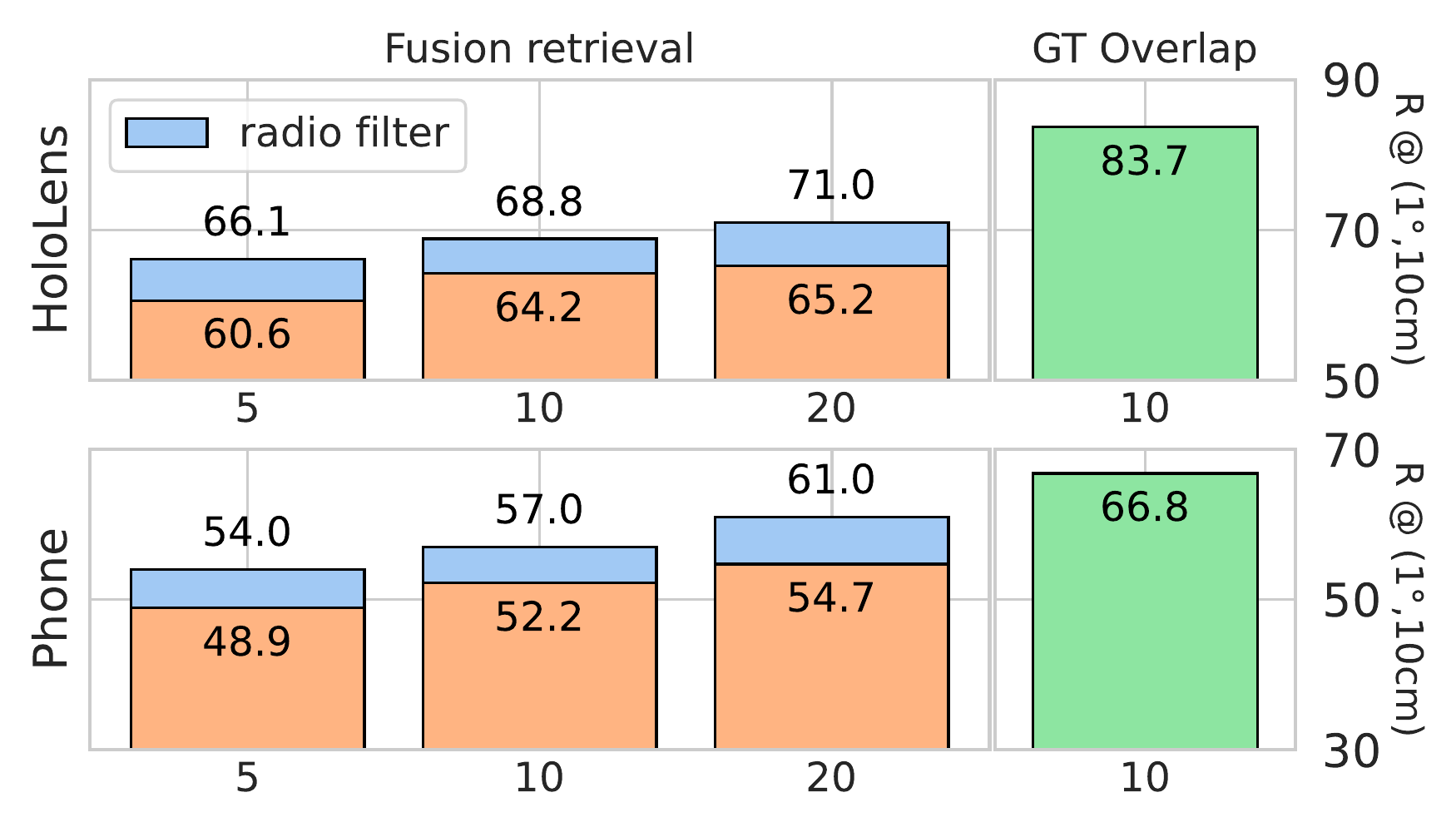}
\end{minipage}
\vspace{0.5mm}
\caption{%
\textbf{Left: single-frame localization.}
We report the recall at ($1^\circ, 10$cm)/($5^\circ, 1$m) for baselines representative of the state of the art.
Our dataset is challenging while most others are saturated.
There is a clear progress from SIFT but also large room for improvement.
\textbf{Right: localization with radio signals.}
Increasing the number \{5, 10, 20\} of retrieved images increases the localization recall at ($1^\circ, 10$cm).
The best-performing visual retrieval (Fusion, orange) is however far worse than the GT overlap.
Filtering with radio signals (blue) improves the performance in all settings.
}%
\label{tab:baselines+radio}
\end{table}

\PAR{Leveraging radio signals:}\label{sec:eval:single-frame-radio}
In this experiment, we show that radio signals can be used to constrain the search space for image retrieval.
This has two main benefits: 1) it reduces the risk of incorrectly considering visual aliases, and 2) it lowers the compute requirements by reducing that numbers of images that need to be retrieved and matched.
We implement this filtering as follows.
We first split the scene into a sparse 3D grid considering only voxels containing at least one mapping frame.
For each frame, we gather all radio signals in a $\pm2$s window and associate them to the corresponding voxel.
If the same endpoint is observed multiple times in a given voxel, we average the received signal strengths (RSSI) in dBm.
For a query frame, we similarly aggregate signals over the past 2s and rank voxels by their L2 distance between RSSIs, considering those with at least one common endpoint.
We thus restrict image retrieval to $2.5\%$ of the map.

\Cref{tab:baselines+radio} (right) shows that radio filtering always improves the localization accuracy over vanilla vision-only retrieval, irrespective of how many images are matches.
The upper bound based on the GT overlap, defined in \cref{sec:method:ref}, shows that there is still much room for improvement for both image and radio retrieval.
As the GT overlap baseline is far from the perfect 100\% recall, frame-to-frame matching and pose estimation have also much room to improve.

\PAR{Varying field-of-view:}
We study the impact of the FOV of the HoloLens 2 device via two configurations:
1) Each camera in a rig is seen as a single-frame and localized using LO-RANSAC + P3P.
2) We consider all four cameras in a frame and localize them together using the generalized solver GP3P.
With fusion retrieval, SuperPoint, and SuperGlue, single images (1) only achieve 45.6\%~/~61.3\% recall, 
while using rigs (2) yields 64.2\%~/~77.4\%~(\cref{tab:baselines+radio}).
Rig localization is thus highly beneficial, especially in hard cases where single cameras face texture-less areas, such as the ground and walls.

\PAR{Mapping modality:}
We study whether the high-quality lidar mesh can be used for localization.
We consider two approaches to obtain a sparse 3D point cloud:
1) By triangulating sparse visual correspondences across multiple views.
2) By lifting 2D keypoints in reference images to 3D by tracing rays through the mesh.
Lifting can leverage dense correspondences, which cannot be efficiently triangulated with conventional multi-view geometry.
We thus compare 1) and 2) with SuperGlue to 2) with LoFTR~\cite{sun2021loftr}, a state-of-the-art dense matcher.
The results in \cref{tab:mapping+tri} (right) show that the mesh brings some improvements.
Points could also be lifted by dense depth from multi-view stereo.
We however did not obtain satisfactory results with a state-of-the-art approach~\cite{wang2020patchmatchnet} as it cannot handle very sparse mapping images.

\PAR{Mapping scenario:}
We study the accuracy of localization against maps built from different types of images: 1) crowd-sourced, dense AR sequences; 2) curated, sparser HD 360 images from the NavVis device; 3) a combination of the two.
The results are summarized in \cref{tab:mapping+tri} (left), showing that the mapping scenario has a large impact on the final numbers.
On the other hand, image pair selection for mapping matters little.
Crowd-sourcing and manual scans can complement each other well to address an imperfect scene coverage.
We hope that future work can close the gap between the scenarios to achieve better metrics from crowd-sourced data without curation.

\begin{table*}[t]
\centering
\begin{minipage}{0.57\linewidth}
\scriptsize{\setlength\tabcolsep{2pt}
\begin{tabular}{llcccc}
    \toprule
    \multicolumn{2}{c}{Mapping images $\rightarrow$} & \multicolumn{2}{c}{HL2 + Phone} & HD 360 & Both \\
    \cmidrule(lr){3-4} \cmidrule(lr){5-5} \cmidrule(lr){6-6}
    \multicolumn{2}{c}{Image pairs from $\rightarrow$} & \multirowcell{2}[-0.0cm]{Retrieval\\+ Poses} & \multirowcell{2}[-0.0cm]{GT\\overlap}
    & \multirowcell{2}[-0.0cm]{Retrieval\\+ Poses} &
    \multirowcell{2}[-0.0cm]{Retrieval\\+ Poses} \\
    Matching & Device\\
    \midrule
    \multirow{2}{*}{SP + SG} & HL2 & 64.2 / 77.4 & 64.2 / 77.3 & 70.1 / 83.6 & 64.1 / 77.5\\
    & Phone & 52.2 / 65.8 & 52.9 / 66.3 & 47.4 / 64.9 & 60.6 / 72.1\\
    \bottomrule
\end{tabular}
\setlength\tabcolsep{3pt}}%
\end{minipage}%
\hspace{0.03\linewidth}%
\begin{minipage}{0.399\linewidth}
\includegraphics[width=1.0\textwidth]{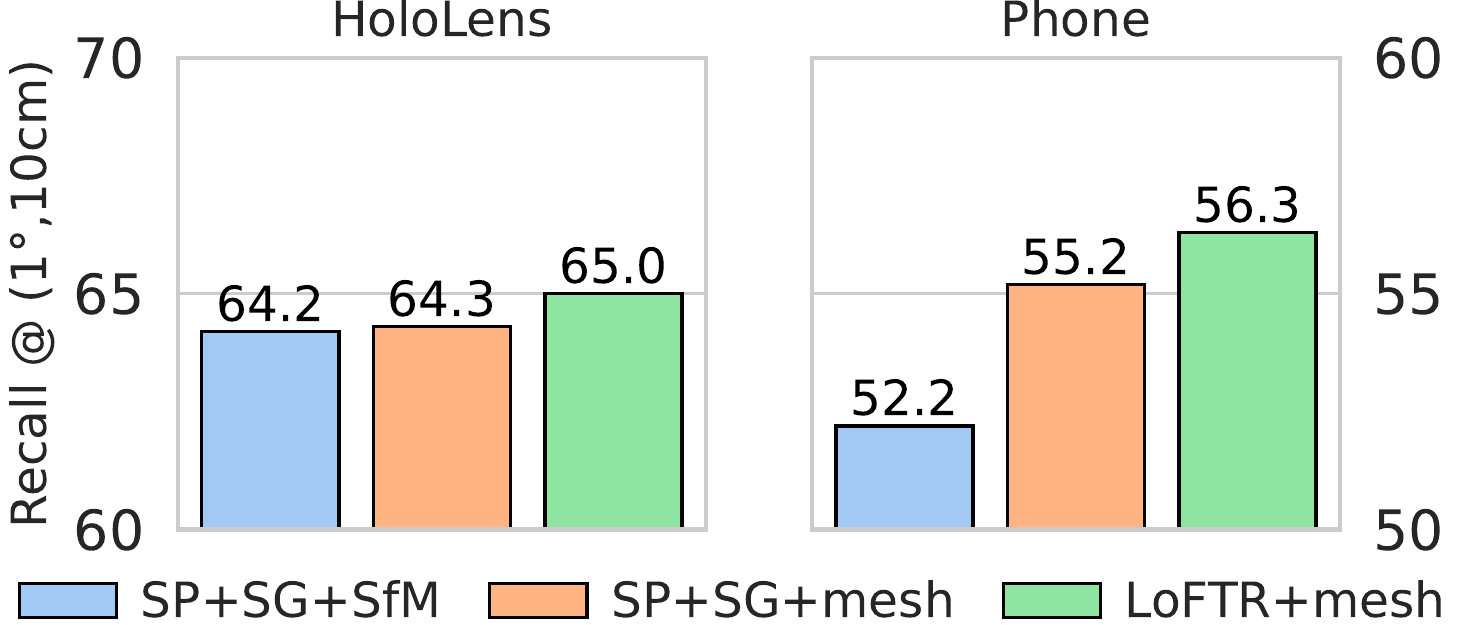}%
\end{minipage}%

\vspace{1.5mm}
\caption{%
\textbf{Impact of mapping.}
\textbf{Left: Scenarios.}
Building the map with HD 360 images from NavVis scanners, instead of or with dense AR sequences, does not consistently boost the performance as they are usually sparser, do not fully cover each location, and have different characteristics than AR images.
\textbf{Right: Modalities.}
Lifting 2D points to 3D using the lidar mesh instead of triangulating with SfM is beneficial.
This can also leverage dense matching, \eg with LoFTR.
}%
\label{tab:mapping+tri}%
\end{table*}

\subsection{Sequence localization}\label{sec:eval:chunk}

In this section, inspired by typical AR use cases, we consider the problem of sequence localization. The task is to align multiple consecutive frames using sensor data aggregated over short time intervals.
Our baseline for this task is based on the ground-truthing pipeline and has as such relatively high compute requirements.
However, we are primarily interested in demonstrating the potential performance gains by leveraging multiple frames.
First, we run image retrieval and single-frame localization, followed by a first PGO with tracking and localization poses.
Then, we do a second localization with retrieval guided by the poses of the first PGO, followed by a second PGO.
Finally, we run a pose refinement by considering reprojections to query frames and tracking cost.
We can also use radio signals to restrict image retrieval throughout the pipeline.
As previously, we consider the localization recall but only of the last frame in each sequence, which is the one that influences the current AR user experience in a real-time scenario.

\begin{figure}[t]
\centering
\begin{minipage}{0.66\linewidth}
\includegraphics[width=\linewidth]{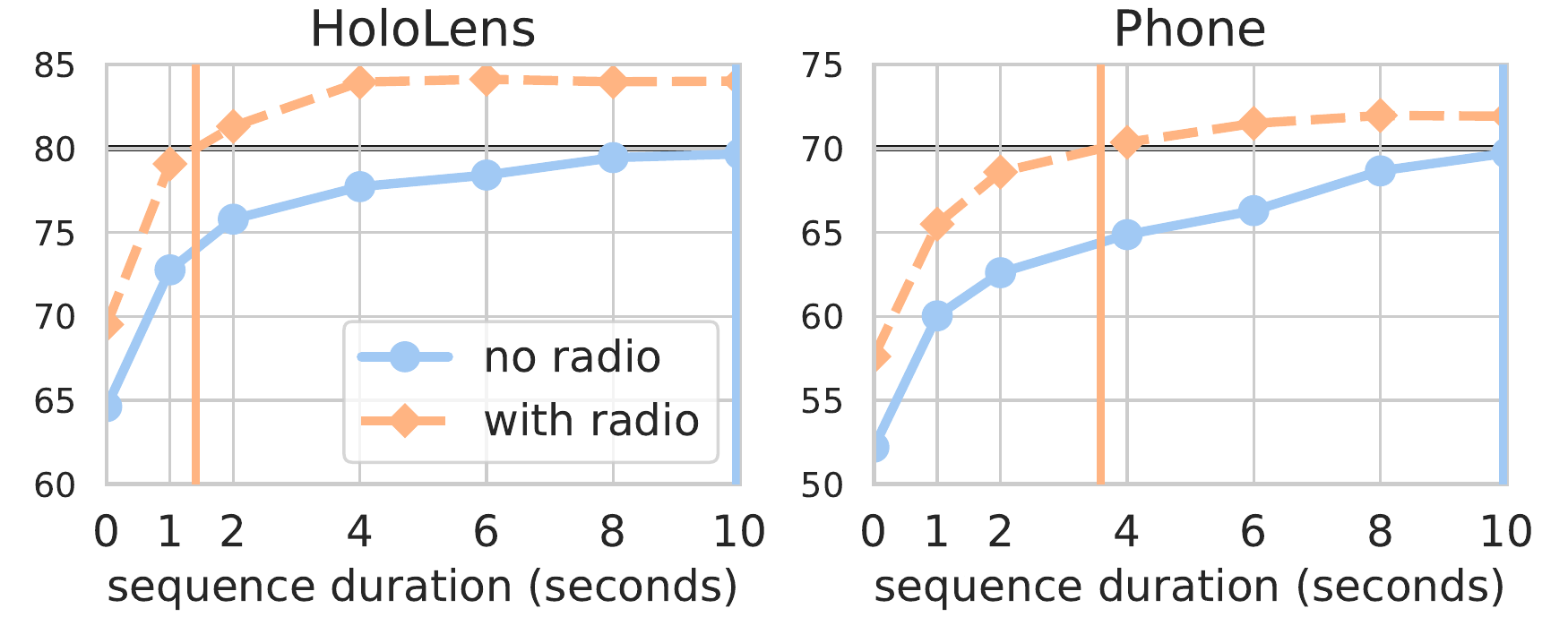}%
\end{minipage}%
\begin{minipage}{0.33\linewidth}
\centering
\scriptsize{\setlength\tabcolsep{3pt}
\begin{tabular}{lcccc}
    \toprule
    \multirow{2}{*}{Device} & \multirow{2}{*}{Radios?} & \multicolumn{3}{c}{TTR@X\%}\\
    \cmidrule(lr){3-5}
    && 70\% & 80\% & 90\%\\
    \midrule
    \multirow{2}{*}{HL2} & \sxmark & \green{<1s} & \red{>10s} & \red{>10s}\\
    & \scmark & \green{<1s} & 1.40s & \red{>10s}\\
    \midrule
    \multirow{2}{*}{Phone} & \sxmark & \red{>10s} & \red{>10s} & \red{>10s}\\
    & \scmark & 3.58s & \red{>10s} & \red{>10s}\\
    \bottomrule
\end{tabular}
\setlength\tabcolsep{3pt}}
\end{minipage}
\vspace{1mm}
\caption{%
\textbf{Sequence localization.}
We report the localization recall at ($1^\circ, 10$cm) of SuperPoint features with SuperGlue matcher as we increase the duration of each sequence.
The pipeline leverages both on-device tracking and absolute localization, as vision-only (solid) or combined with radio signals (dashed).
We show the time-to-recall (TTR) at 80\% for HL2 and at 70\% for phone queries.
Using radio signals reduces the TTR from over 10s to 1.40s and 3.58s, respectively.
}
    \label{fig:chunks}
\end{figure}

We evaluate various query durations and introduce the \emph{time-to-recall} metric as the sequence length (time) required to successfully localize X\% (recall) of the queries within (1$^\circ$, 10cm), or, in short, TTR@X\%.
Localization algorithms should aim to minimize this metric to render retrieved content as quickly as possible after starting an AR experience.
\Cref{fig:chunks} reports the results averaged over all locations.
While the performance of current methods is not satisfactory yet to achieve a TTR@90\% under 10 seconds, using sequence localization leads to significant gains of 20\%.
The radio signals improve the performance in particular with shorter sequences and thus effectively reduce the time-to-recall.

\section{Conclusion}

LaMAR is the first benchmark that faithfully captures the challenges and opportunities of AR for visual localization and mapping.
We first identified several key limitations of current benchmarks that make them unrealistic for AR.
To address these limitations, we developed a new ground-truthing pipeline to accurately and robustly register AR sensor streams in large and diverse scenes aided by laser scans without any manual labelling or custom infrastructure.
With this new benchmark, initially covering 3 large locations, we revisited the traditional academic setup and showed a large performance gap for existing state-of-the-art methods when evaluated using more realistic and challenging data.

We implemented simple yet representative baselines to take advantage of the AR-specific setup and we presented new insights that pave promising avenues for future works.
We showed the large potential of leveraging other sensor modalities like radio signals, depth, or query sequences instead of single images.
We also hope to direct the attention of the community towards improving map representations for crowd-sourced data and towards considering the time-to-recall metric, which is currently largely ignored.
We publicly release at \href{https://lamar.ethz.ch/}{\texttt{lamar.ethz.ch}} the complete LaMAR dataset, our ground-truthing pipeline, and the implementation of all baselines.
The evaluation server and public leaderboard facilitates the benchmarking of new approaches to keep track of the state of the art.
We hope this will spark future research addressing the challenges of AR.

\PAR{Acknowledgements.} 
LaMAR would not have been possible without the hard work and contributions of %
Gabriela Evrova, 
Silvano Galliani, 
Michael Baumgartner, 
Cedric Cagniart, 
Jeffrey Delmerico, 
Jonas Hein, 
Dawid Jeczmionek, 
Mirlan Karimov, 
Maximilian Mews,
Patrick Misteli, 
Juan Nieto,
Sònia Batllori Pallarès, 
R\'emi Pautrat, 
Songyou Peng, 
Iago Suarez,
Rui Wang, 
Jeremy Wanner, 
Silvan Weder 
and our colleagues in \mbox{CVG at ETH Zurich} and the wider Microsoft Mixed Reality \& AI team.

\fi

\ifincludesupp
\appendix

\section*{Appendix}

\section{Visualizations}
\label{sec:visuals}
\PAR{Diversity of devices:} 
We show in \cref{fig:supp:samples} some samples of images captured in the HGE location.
NavVis and phone images are colored while HoloLens2 images are grayscale.
NavVis images are always perfectly upright, while the viewpoint and height of HoloLens2 and phone images varies significantly.
Despite the automatic exposure, phone images easily appear dark in night-time low-light conditions.

\PAR{Diversity of environments:} 
We show an extensive overview of the three locations CAB, HGE, and LIN in Figures~\ref{fig:supp:CAB}, \ref{fig:supp:HGE}, and \ref{fig:supp:LIN}, respectively.
In each image, we show a rendering of the lidar mesh along with the ground truth trajectories of a few sequences.

\begin{figure}[!b]
    \centering
    \includegraphics[width=\textwidth]{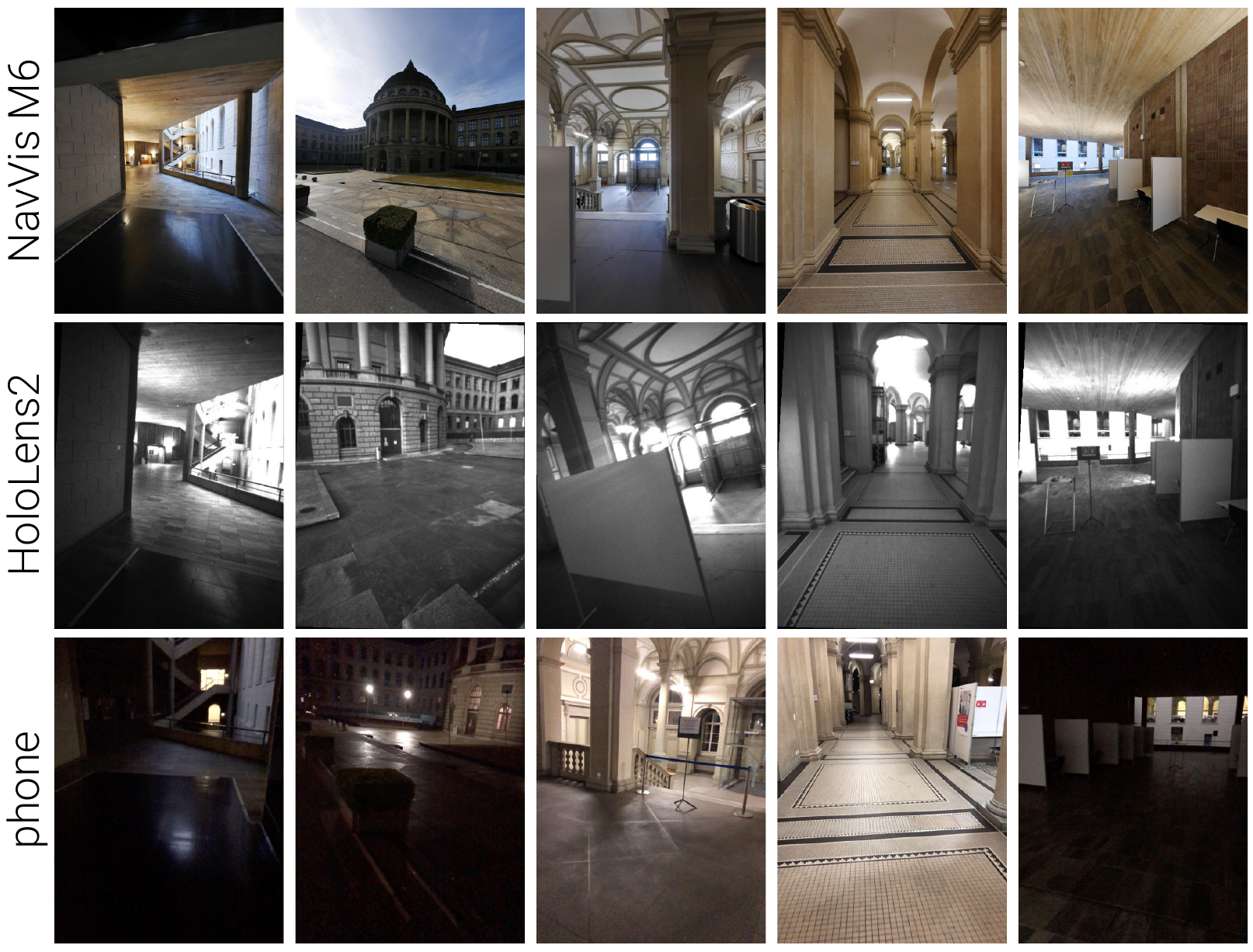}
    \caption{\textbf{Sample of images from the different devices:} NavVis M6, HoloLens2, phone.
    Each column shows a different scene of the HGE location with large illumination changes.
    }%
    \label{fig:supp:samples}%
\end{figure}

\begin{figure}[p]
    \centering
    \input{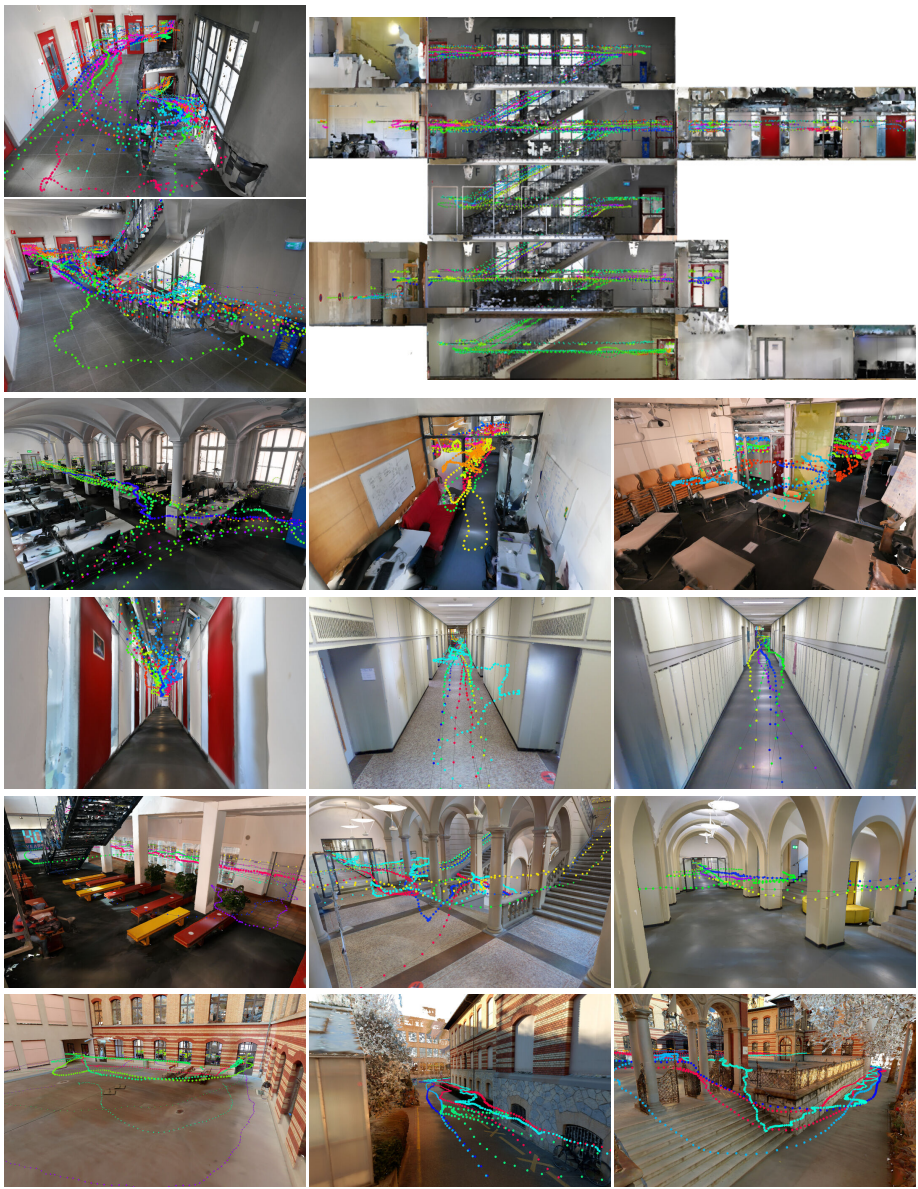}
    \vspace{2mm}
    \caption{\textbf{The CAB location} features 1-2) a staircase spanning 5 similar-looking floors, 3) large and small offices and meeting rooms, 4) long corridors, 5) large halls, and 6) outdoor areas with repeated structures.
    This location includes the \emph{Facade}, \emph{Courtyard}, \emph{Lounge}, \emph{Old Computer}, \emph{Storage Room}, and \emph{Office} scenes of the ETH3D~\cite{schops2017multi} dataset and is thus much larger than each of them.
    }%
    \label{fig:supp:CAB}%
\end{figure}

\begin{figure}[!ht]
    \centering
    \input{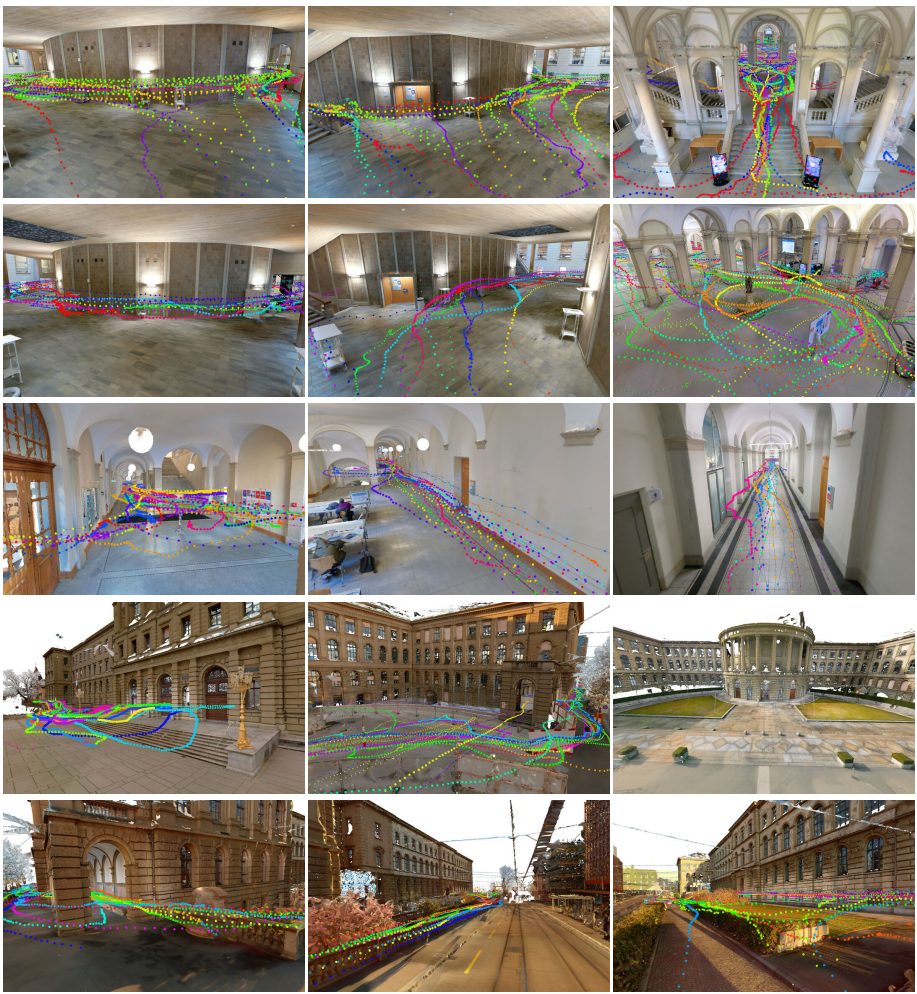}
    \vspace{2mm}
    \caption{\textbf{The HGE location} features a highly-symmetric building with 1-2) hallways, 3) long corridors, 4) two esplanades, and 5) a section of sidewalk.
    This location includes the \emph{Relief}, \emph{Door}, and \emph{Statue} scenes of the ETH3D~\cite{schops2017multi} dataset.
    }%
    \label{fig:supp:HGE}
    \vspace{1cm}
\end{figure}

\begin{figure}[ht]
    \centering
    \input{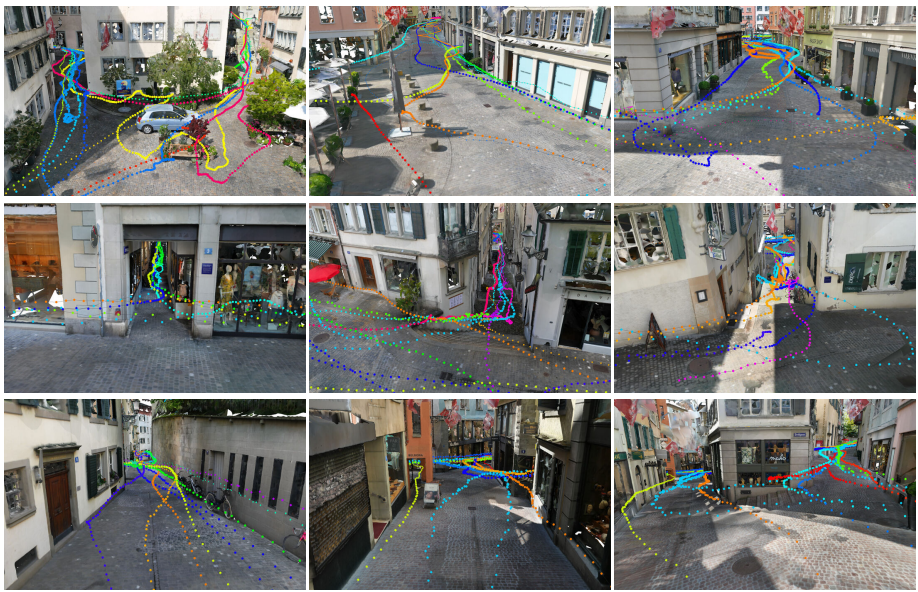}
    \vspace{2mm}
    \caption{\textbf{The LIN location} features large outdoor open spaces (top row), narrow passages with stairs (middle row), and both residential and commercial street-level facades.
    }%
    \label{fig:supp:LIN}
    \vspace{1cm}
\end{figure}

\PAR{Long-term changes:}
Because spaces are actively used and managed, they undergo significant appearance and structural changes over the year-long data recording.
This is captured by the laser scans, which are aligned based on elements that do not change, such as the structure of the buildings.
We show in \cref{fig:supp:changes} a visual comparison between scans captured at different points in time.

\begin{figure}[p]
    \centering
    \input{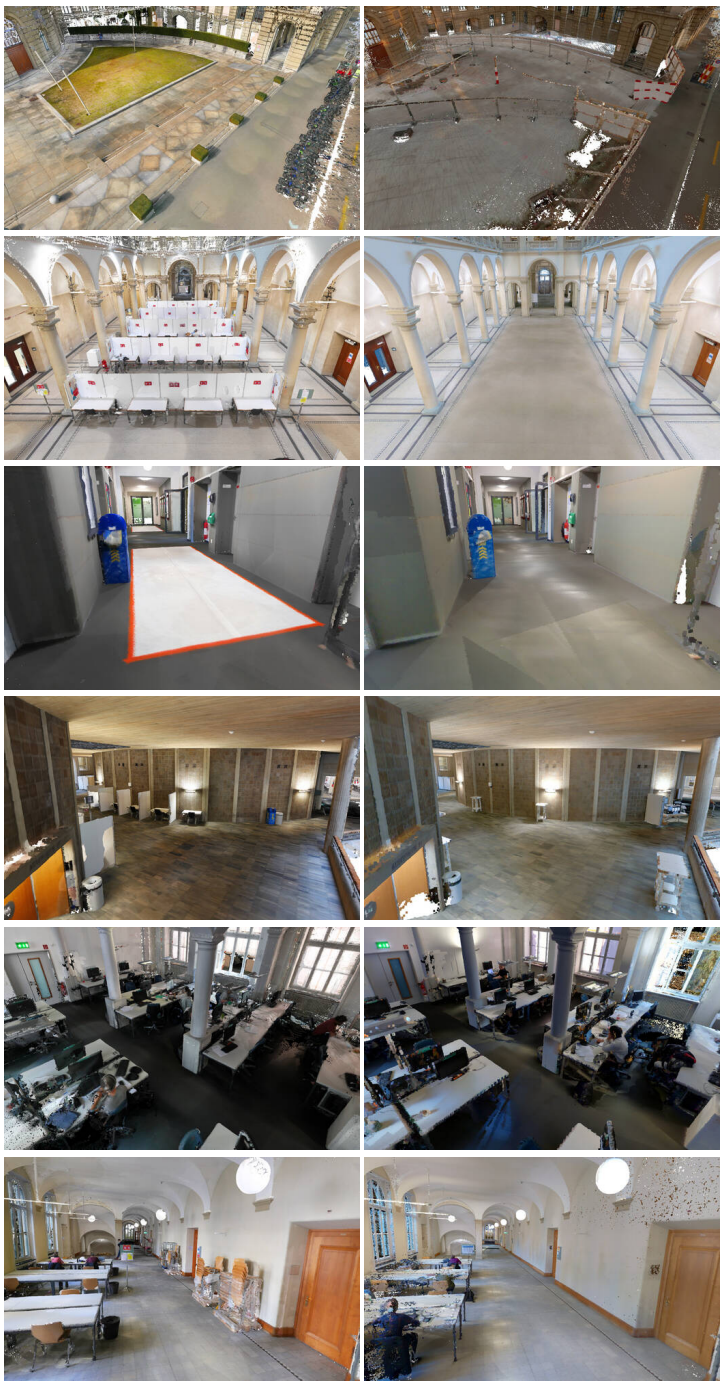}
    \vspace{1mm}
    \caption{\textbf{Long-term structural changes.}
    Lidar point clouds captured over a year reveal the geometric changes that spaces undergo at different time scales:
    1) very rarely (construction work),
    2-4) sparsely (displacement of furniture), or even
    5-6) daily due to regular usage (people, objects).
    }%
    \label{fig:supp:changes}%
\end{figure}

\newpage
\section{Uncertainties of the ground truth}
\label{sec:uncertainties}
We show overhead maps and histograms of uncertainties for all scenes in \cref{fig:supp:uncertainties}.
We also additional rendering comparisons in \cref{fig:renderings}.
Since we do not use the mesh for any color-accurate tasks (e.g., photo-metric alignment), we use a simple vertex-coloring based on the NavVis colored lidar pointcloud.
The renderings are therefore not realistic but nevertheless allow an inspection the final alignment.
The proposed ground-truthing pipeline yields poses that allow pixel-accurate rendering.

\begin{figure}[p]
    \centering
    \input{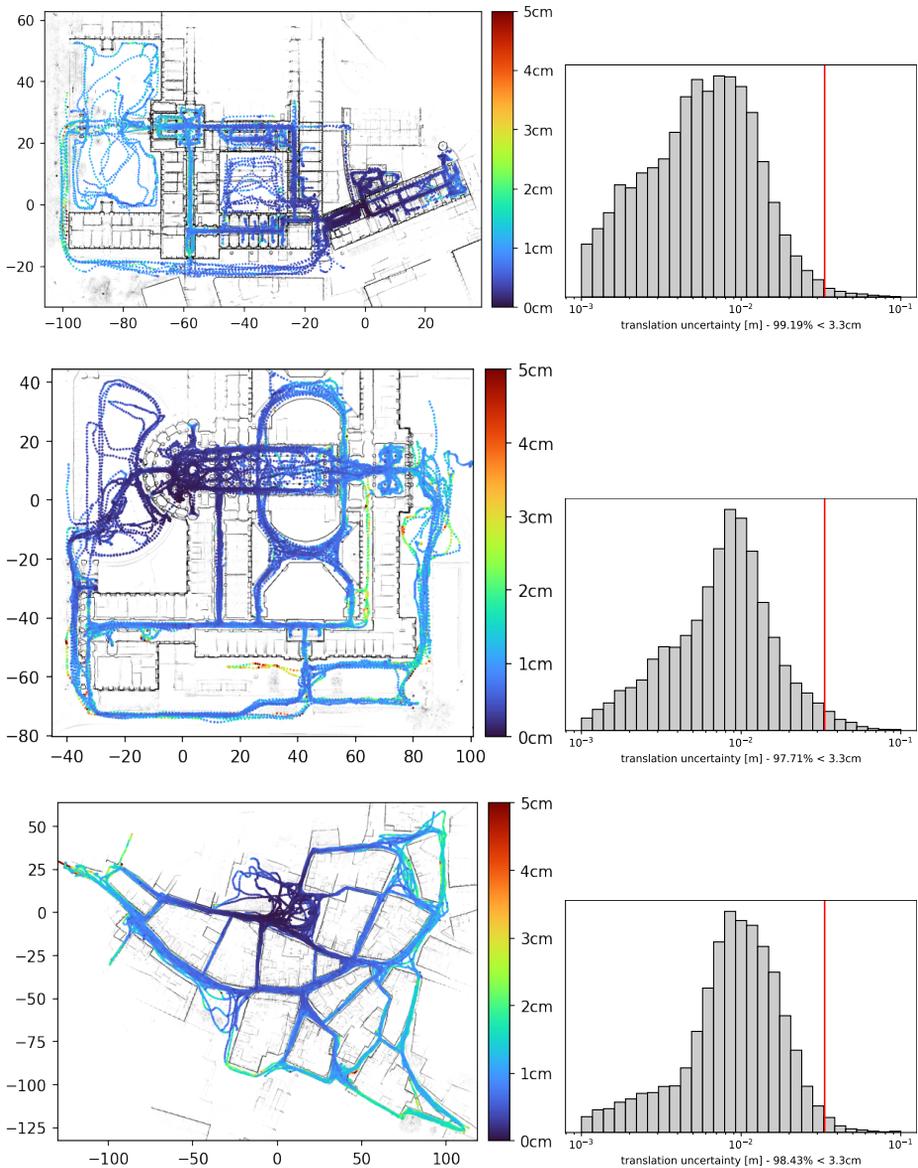}
    \vspace{4mm}
    \caption{\textbf{Translation uncertainties of the ground truth camera centers} for the CAB (top), LIN (middle) and HGE (bottom) scenes.
    Left: The overhead map shows that the uncertainties are larger in areas that are not well covered by the 3D scanners or where the scene is further away from the camera, such as in long corridors and large outdoor space.
    Right: The histogram of uncertainties shows that most images have an uncertainty far lower than $\sigma_t{=}3.33\text{cm}$.
    }%
    \label{fig:supp:uncertainties}%
\end{figure}

\begin{figure}[p]
    \centering
    \includegraphics[width=.30\textwidth]{figures/qualitative_renderings/116376895_hl_2022-01-18-12-58-38-108.000_hetlf.jpg}~
    \includegraphics[width=.30\textwidth]{figures/qualitative_renderings/2714445675_hl_2021-06-02-11-59-31-495.001_hetlf.jpg}~
    \includegraphics[width=.30\textwidth]{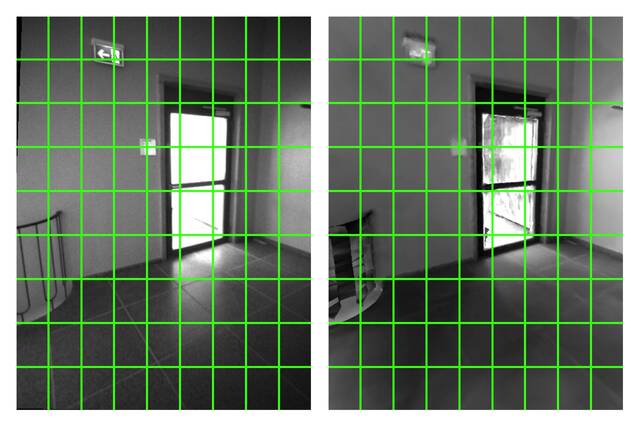}\\
    \includegraphics[width=.30\textwidth]{figures/qualitative_renderings/131582775_hl_2022-01-18-12-58-38-108.000_hetlf.jpg}~
    \includegraphics[width=.30\textwidth]{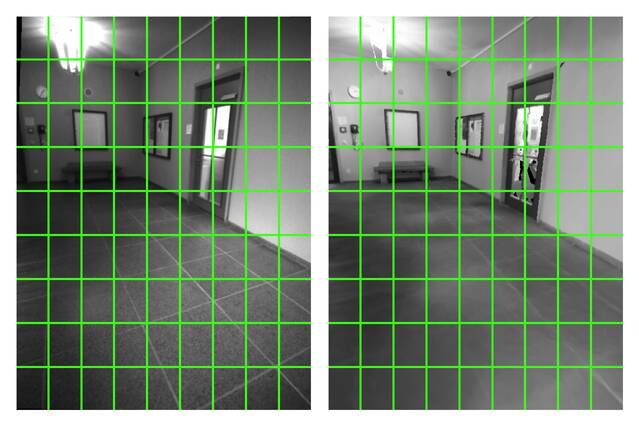}~
    \includegraphics[width=.30\textwidth]{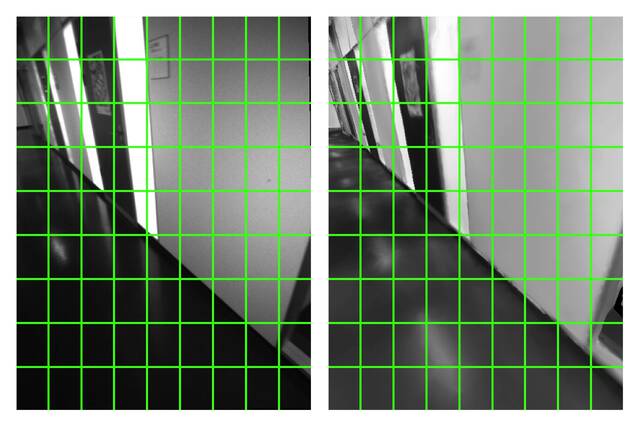}\\
    \includegraphics[width=.30\textwidth]{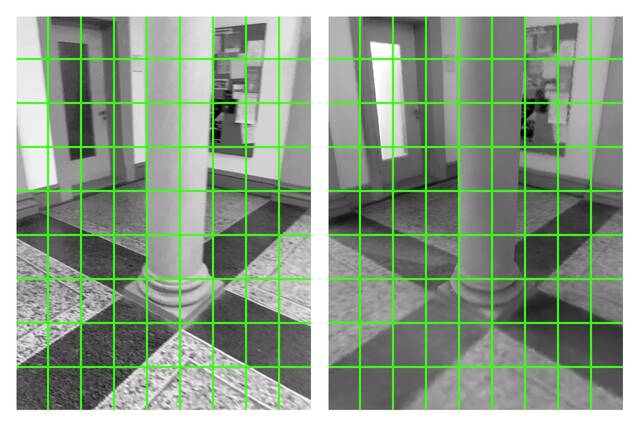}~  \includegraphics[width=.30\textwidth]{figures/qualitative_renderings/6493927726_ios_2022-02-27_18.05.07_000_cam_phone_6493927726.jpg}~    \includegraphics[width=.30\textwidth]{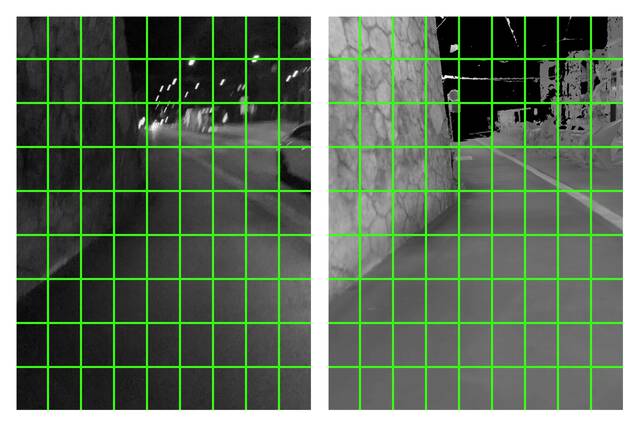}\\ \includegraphics[width=.30\textwidth]{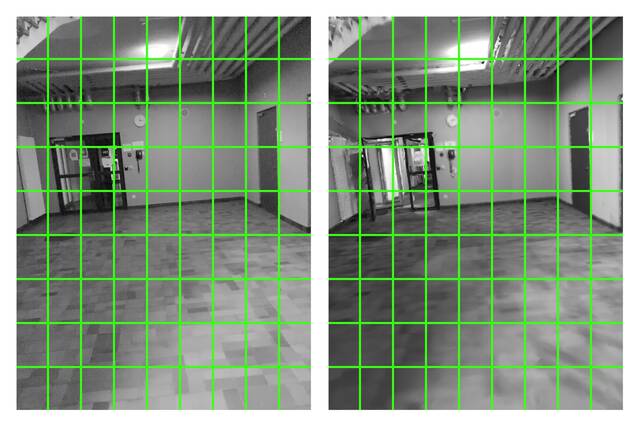}~    \includegraphics[width=.30\textwidth]{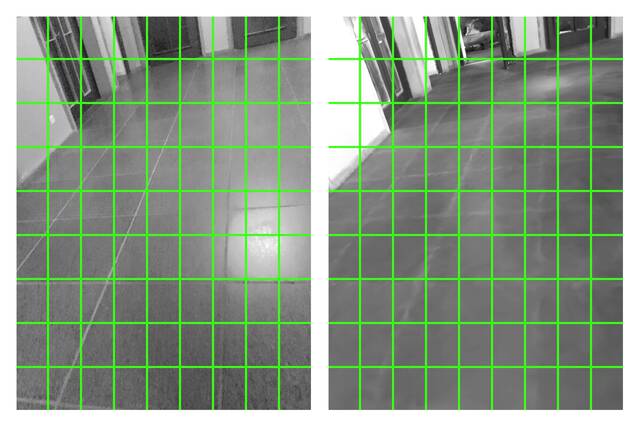}~
    \includegraphics[width=.30\textwidth]{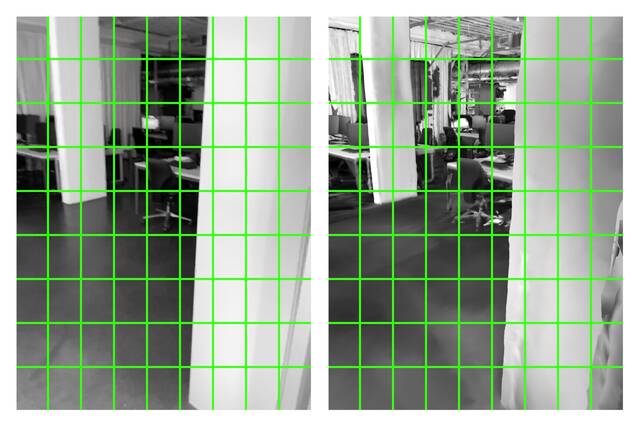}\\
    \includegraphics[width=.15\textwidth]{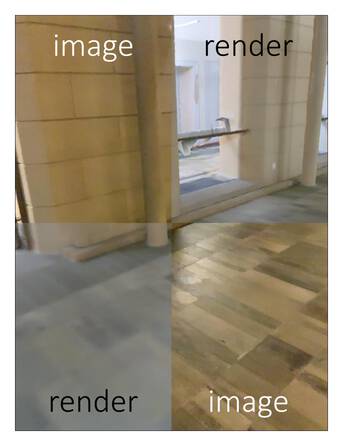}~
    \includegraphics[width=.15\textwidth]{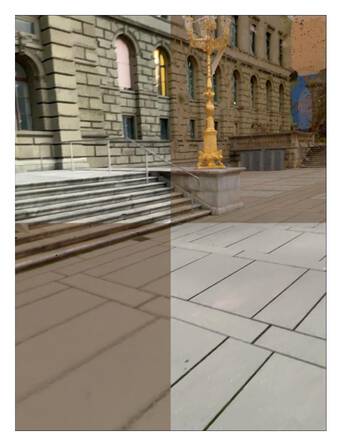}~
    \includegraphics[width=.15\textwidth]{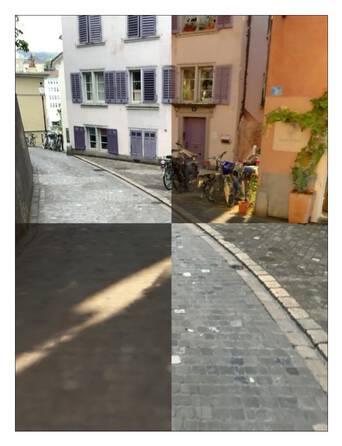}~
    \includegraphics[width=.15\textwidth]{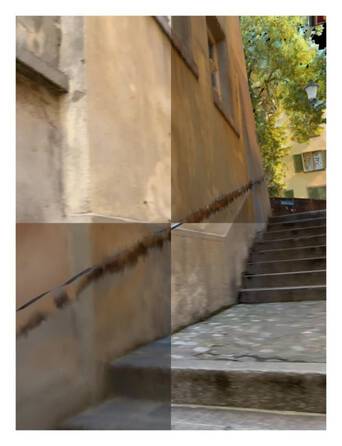}~
    \includegraphics[width=.15\textwidth]{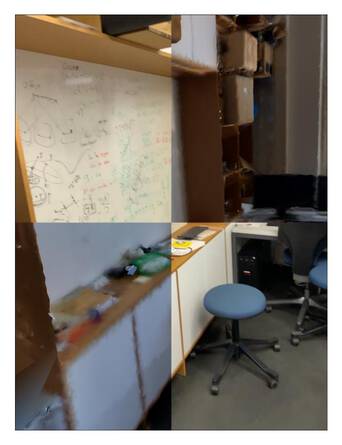}~
    \includegraphics[width=.15\textwidth]{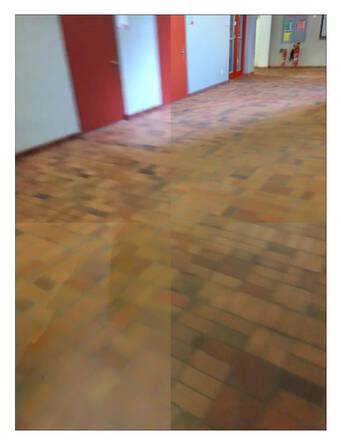}\\
    \includegraphics[width=.15\textwidth]{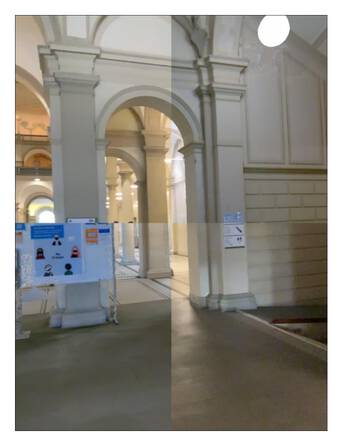}~
    \includegraphics[width=.15\textwidth]{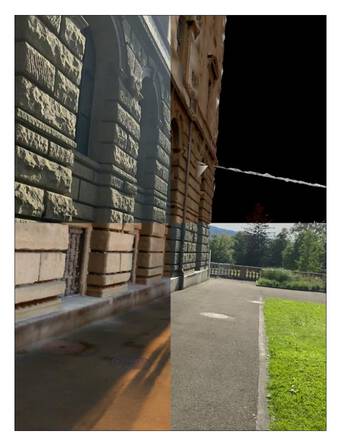}~
    \includegraphics[width=.15\textwidth]{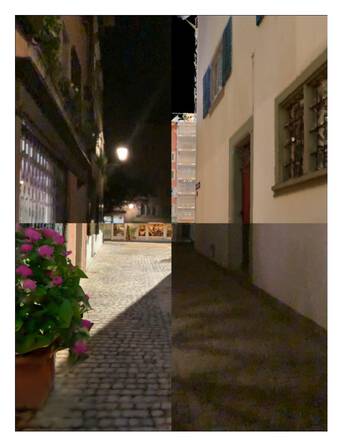}~
    \includegraphics[width=.15\textwidth]{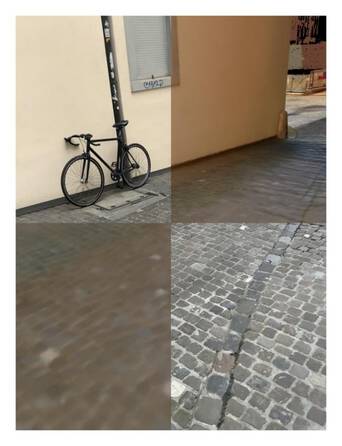}~
    \includegraphics[width=.15\textwidth]{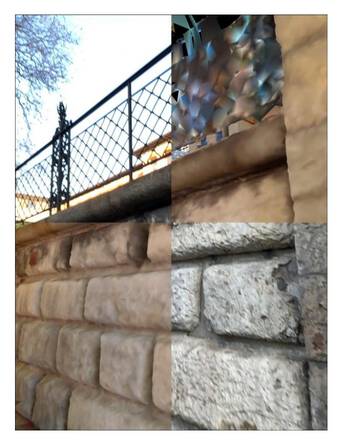}~
    \includegraphics[width=.15\textwidth]{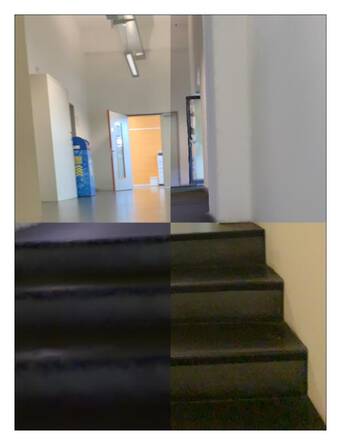}\\
    \vspace{1mm}
    \caption{{\bf Qualitative renderings from the mesh.}
    Using the estimated ground-truth poses, we render images from the vertex-colored mesh (right) and compare them to the originals (left).
    The first two rows show six HoloLens 2 images while the next two show six phone images.
    We overlay a regular grid to facilitate the comparison.
    The bottom rows shows 2x2 mosaics alternating between originals (top-left, bottom-right) and renders (top-right, bottom-left).
    Best viewed when zoomed in.
    }
    \label{fig:renderings}
\end{figure}

\newpage
\section{Selection of mapping and query sequences}
\label{sec:map-query-split}
We now describe in more details the algorithm that automatically selects mapping and query sequences, whose distributions are shown in \cref{fig:supp:qualitymaps}.

The coverage $C(i,j)_k$ is a boolean that indicates whether the image $k$ of sequence $i$ shares sufficient covisibility with at least one image is sequence $j$.
Here two images are deemed covisible if they co-observe a sufficient number of 3D points in the final, full SfM sparse model~\cite{radenovic2018fine} or according to the ground truth mesh-based visual overlap.
The coverage of sequence $i$ with a set of other sequences $\mathcal{S} = \{j\}$ is the ratio of images in $i$ that are covered by at least one image in $\mathcal{S}$: 
\begin{equation}
C(i,\mathcal{S}) = \frac{1}{|i|}\sum_{k\in i} \bigcap_{j \in \mathcal{S}} C(i,j)_k
\end{equation}

\begin{figure}[p]
    \centering
    \input{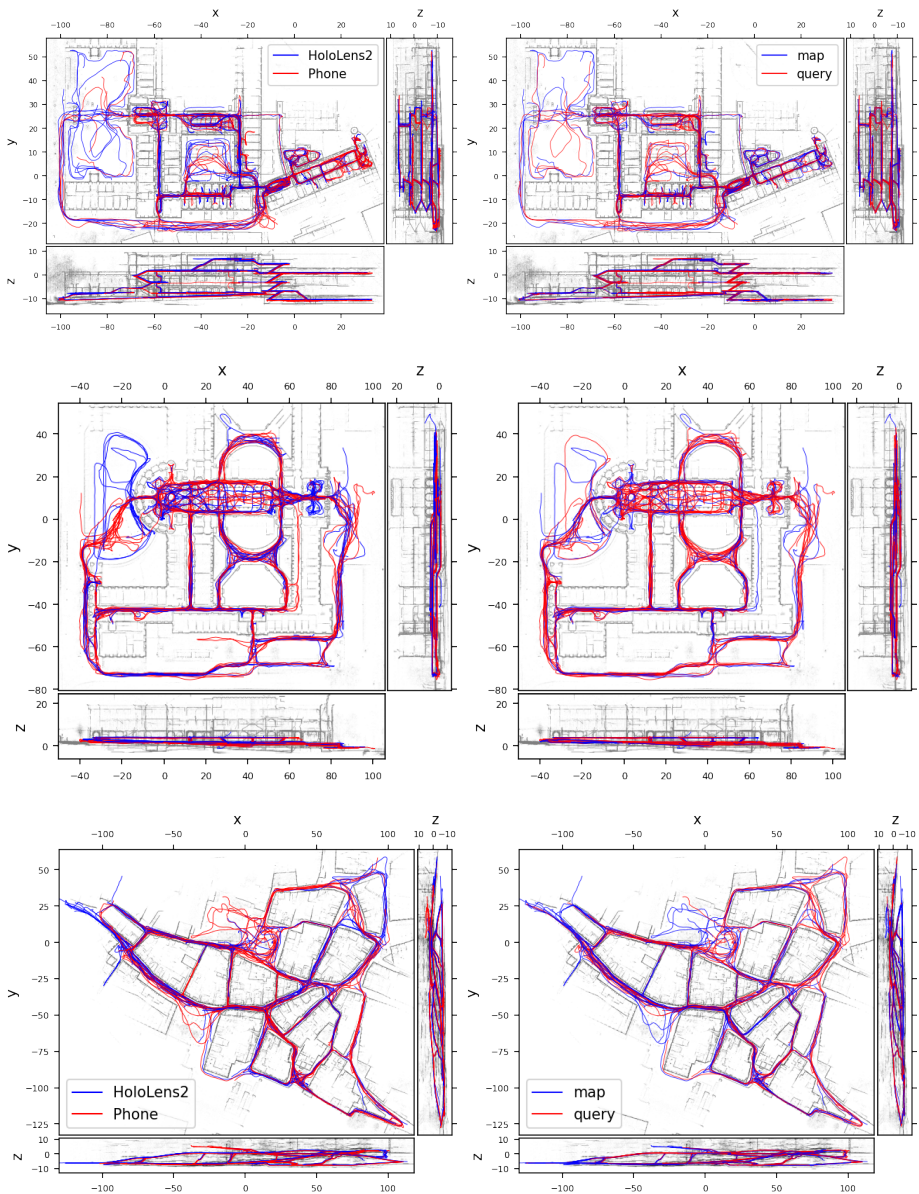}
    \vspace{4mm}
    \caption{\textbf{Spatial distribution of AR sequences} for the CAB (top), HGE (middle), and LIN (bottom) locations.
    We show the ground truth trajectories overlaid on the lidar point clouds along 3 orthogonal directions.
    All axes are in meters and $z$ is aligned with the gravity.
    Left: Types of AR devices among all registered sequences.
    Right: Map and query sequences selected for evaluation.
    CAB spans multiple floors while HGE and LIN are mostly 2D but include a range of ground heights.
    The space is well covered by both types of devices and sequences.
    }%
    \label{fig:supp:qualitymaps}%
\end{figure}

We seek to find the set of mapping sequences $\mathcal{M}$ and remaining query sequences $\mathcal{Q} = \mathcal{S}\backslash\mathcal{M}$ that minimize the coverage between map sequences while ensuring that each query is sufficiently covered by the map:
\begin{align}
\begin{split}
    \mathcal{M}^* = \argmin \frac{1}{|\mathcal{M}|} \sum_{i \in\mathcal{M}} C(i,\mathcal{M}\backslash\{i\}) \\
    \text{such that}\quad C(i,\mathcal{M}) > \tau\quad \forall i \in \mathcal{Q}\enspace,
\end{split}
\end{align}
where $\tau$ is the minimum query coverage.
We ensure that query sequences are out of coverage for at most $t$ consecutive seconds, where $t$ can be tuned to adjust the difficulty of the localization and generally varies from 1 to 5 seconds.
This problem is combinatorial and without exact solution.
We solve it approximately with a depth-first search that iteratively adds new images and checks for the feasibility of the solution.
At each step, we consider the query sequences that are the least covisible with the current map.

\section{Data distribution}
\label{sec:supp:distribution}
We show in \cref{fig:supp:qualitymaps} (left) the spatial distribution to the registered sequences for the two devices types HoloLens2 and phone.

We select a subset of all registered sequences for evaluation and split it into mapping and localization groups according to the algorithm described in \cref{sec:map-query-split}.
The spatial distribution of these groups is shown in \cref{fig:supp:qualitymaps} (right).
We enforce that night-time sequences are not included in the map, which is a realistic assumption for crowd-sourced scenarios.
We do not enforce an equal distribution of device types in either group but observe that this occurs naturally.
For the evaluation, mapping images are sampled at intervals of at most 2.5FPS, 50cm of distance, and 20\degree of rotation. 
This ensures a sufficient covisibility between subsequent frames while reducing the computational cost of creating maps.
The queries are sampled every 1s/1m/20\degree and, for each device type, 1000 poses are selected out of those with sufficiently low uncertainty.

\section{Additional evaluation results}

\subsection{Impact of the condition and environment}
We now investigate the impact of different capture conditions (day vs night) and environment (indoor vs outdoor) of the query images.
Query sequences are labeled as day or night based on the time and date of capture.
We manually annotate overhead maps into indoor and outdoor areas.
We report the results for single-image localization of phone images in \cref{tab:supp:cond-env}.

In regular day-time conditions, outdoor areas exhibit distinctive texture and are thus easier to coarsely localize in than texture-less, repetitive indoor areas.
The scene structure is however generally further away from the camera, so optimizing reprojection errors yields less accurate camera poses.

Indoor scenes generally benefit from artificial light and are thus minimally affected by the night-time drop of natural light.
Outdoor scenes benefit from little artificial light, mostly due to sparse street lighting, and thus widely change in appearance between day and night.
As a result, the localization performance drops to a larger extent outdoors than indoors.

\begin{table}[t]
    \centering
{%
\setlength\tabcolsep{5pt}
\begin{tabular}{cccccc}
    \toprule
    \multirowcell{2}[-0.1cm]{Condition} & \multicolumn{2}{c}{CAB scene} & \multicolumn{2}{c}{HGE scene} & LIN scene\\
    \cmidrule(lr){2-3}
    \cmidrule(lr){4-5}
    \cmidrule(lr){6-6}
    & Indoor & Outdoor & Indoor & Outdoor & Outdoor \\
    \midrule
    day & 66.5 / 74.7 & 73.9 / 88.1 & 52.7 / 65.9 & 43.0 / 64.3 & 71.2 / 82.5\\
    night & 30.3 / 44.8 & 18.8 / 40.6 & 47.9 / 59.4 & 12.1 / 33.6 & 38.6 / 55.6\\
    \bottomrule
\end{tabular}
}
    \vspace{2mm}
    \caption{\textbf{Impact of the condition and environment} on single-image phone localization.
    During the day, localizing indoors can be more accurate (10cm threshold) but less robust (1m threshold) than outdoors due to visual aliasing and a lack of texture.
    Night-time localization is more challenging outdoors than indoors because of a larger drop of illumination.
    }
    \label{tab:supp:cond-env}
\end{table}

\subsection{Additional results on sequence localization}

We run a detailed ablation of the sequence localization on an extended set of queries and report our finding below.

\PAR{Ablation:} We ablate the different parts of our proposed sequence localization pipeline on sequences of 20 seconds.
The localization recall at \{$1^\circ, 10$cm\} and \{$5^\circ, 1$m\} can be seen in \cref{tab:multi-frame-ablation} for both HoloLens 2 and Phone queries.
The initial PGO with tracking and absolute constraints already offers a significant boost in performance compared to single-frame localization.
We notice that the re-localization with image retrieval guided by the PGO poses achieves much better results than the first localization - this points to retrieval being a bottle-neck, not feature matching.
Next, the second PGO is able to leverage the improved absolute constraints and yields better results.
Finally, the pose refinement optimizing reprojection errors while also taking into account tracking constraints further improves the performance, notably at the the tighter threshold.

\begin{table}[t]
    \centering
    \begin{tabular}{cccccccc}
        \toprule
        \multirow{2}{*}{Device} & \multirow{2}{*}{Radios} & \multicolumn{6}{c}{Steps} \\
        \cmidrule(lr){3-8}
        & & Loc. & Init. & PGO1 & Re-loc. & PGO2 & BA\\
        \midrule
        \multirow{2}{*}{HL2} & \sxmark & 66.0 / 79.9 & 66.1 / 92.5 & 71.8 / 92.4 & 74.2 / 88.0 & 74.9 / 92.5   & {\bf 79.3} / {\bf 92.8}  \\
        & \scmark & 67.7 / 82.3 & 66.4 / 94.5 & 74.3 / 94.3 & 76.2 / 90.1 & 76.7 / 94.4 & {\bf 81.6} / {\bf 94.9} \\
        \midrule
        \multirow{2}{*}{Phone} & \sxmark & 54.2 / 65.5 & 52.4 / 88.0 & 62.7 / 87.7 & 61.8 / 77.4 & 66.1 / 88.4 & {\bf 69.0} / {\bf 88.6} \\
        & \scmark & 56.7 / 71.5 & 54.1 / {\bf 90.2} & 64.4 / 89.8 & 63.1 / 79.5 & 66.9 / 90.1 & {\bf 71.0} / {\bf 90.2} \\
         \bottomrule
    \end{tabular}
    \vspace{1mm}
    \caption{{\bf Ablation of the sequence localization.}
    We report recall for the different steps of the sequence localization pipeline for 10s sequences on the CAB location.
    The second localization, guided by the poses of the first PGO, drastically improves over the initial localization, especially when no radio signals are used.
    The final pose refinement optimizing reprojection errors while also taking into account tracking constraints offers a significant boost for the tighter threshold.}
    \label{tab:multi-frame-ablation}
\end{table}

\section{Phone capture application}
We wrote a simple iOS Swift application that saves the data exposed by the ARKit, CoreBluetooth, CoreMotion, and CoreLocation services.
The user can adjust the frame rate prior to capturing.
The interface displays the current input image and the trajectory estimated by ARKit as an AR overlay.
It also displays the amount of free disk space on the device, the recording time, the number of captured frames, and the status of the ARKit tracking and mapping algorithms.
The data storage is optimized such that a single device can capture hours of data without running out of space.
After capture, the data can be inspected on-device and shared over Airdrop or cloud storage.
Screenshots of the user interface are shown in \cref{fig:supp:app}.

\begin{figure}[t]
    \centering
    \begin{minipage}{0.25\textwidth}
        \includegraphics[width=\linewidth]{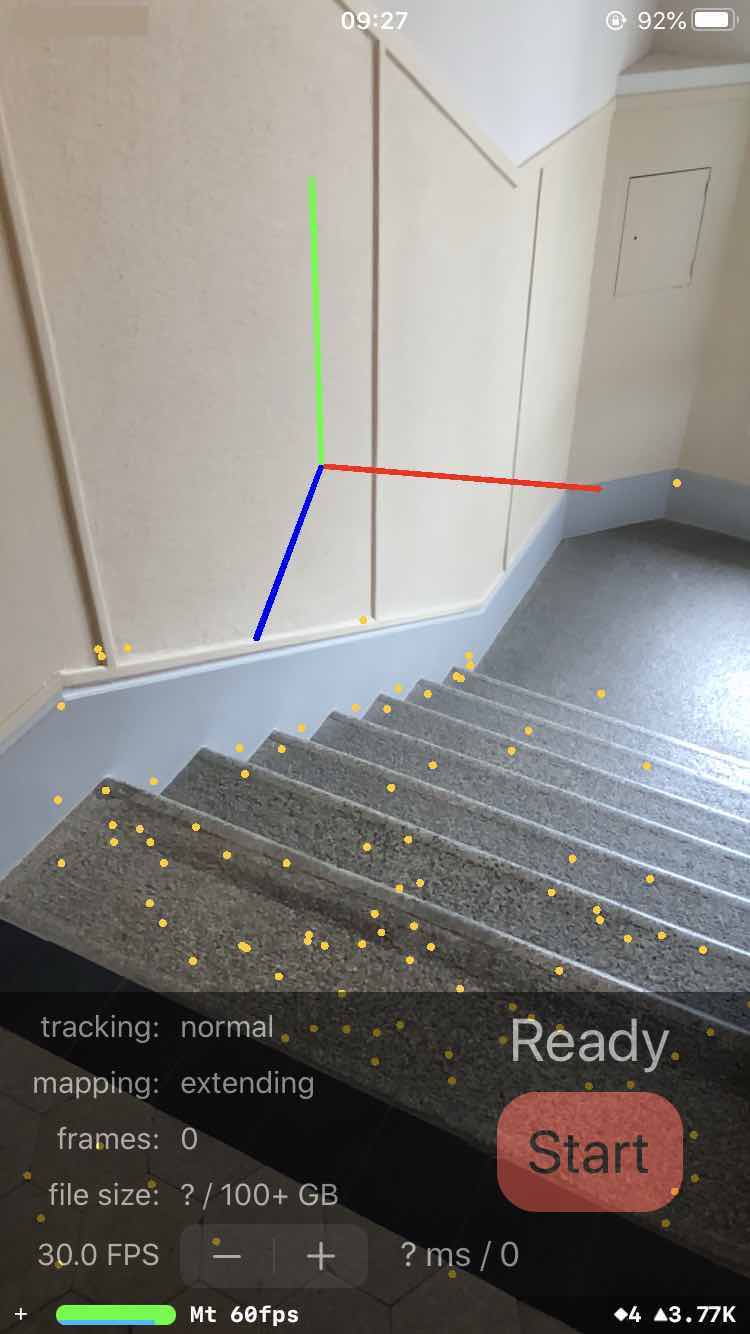}
    \end{minipage}%
    \hspace{1mm}%
    \begin{minipage}{0.25\textwidth}
        \includegraphics[width=\linewidth]{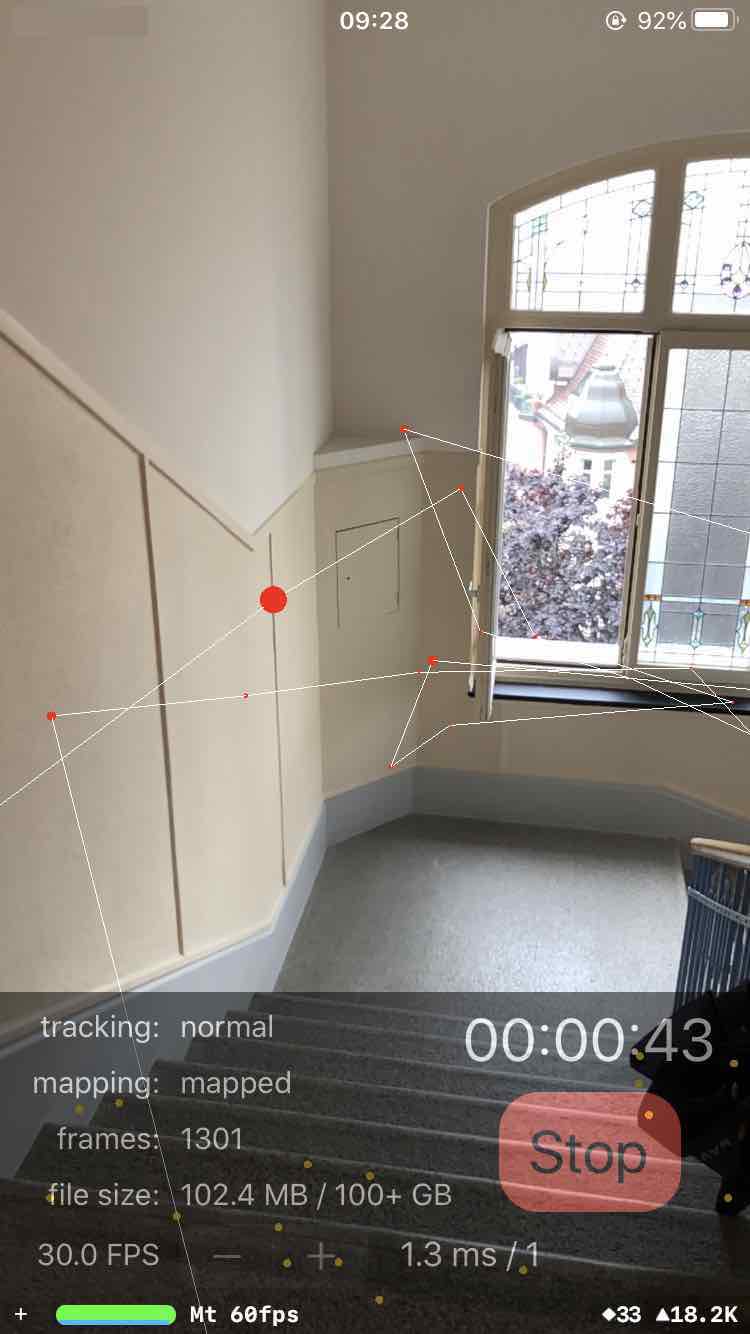}
    \end{minipage}%
    \hspace{1mm}%
    \begin{minipage}{0.25\textwidth}
        \includegraphics[width=\linewidth]{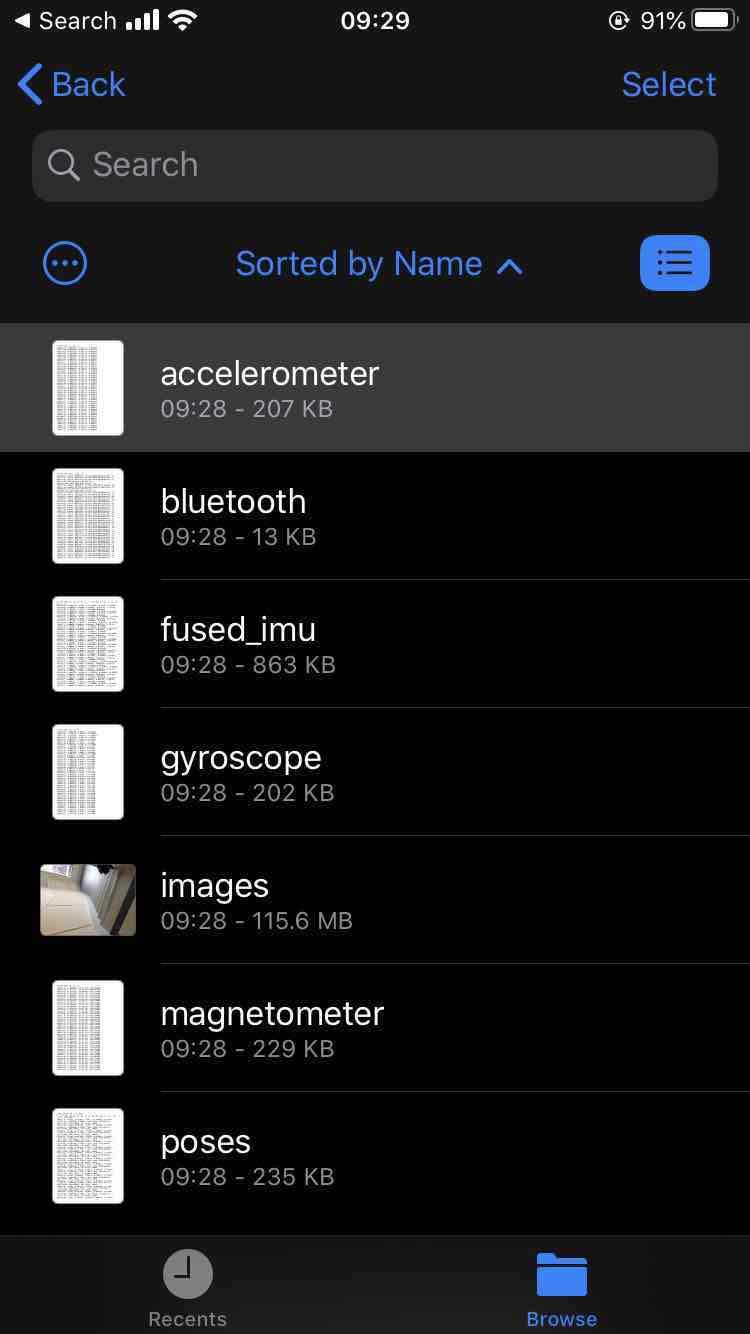}
    \end{minipage}

    \vspace{2mm}
    \caption{\textbf{iOS capture application.}
    }%
    \label{fig:supp:app}%
    \vspace{0.5cm}
\end{figure}

\section{Implementation details}

\PAR{Scan-to-scan alignment:}
The pairwise alignment is initialized by matching the $r{=}5$ most similar images and running 3D-3D RANSAC with a threshold $\tau{=}5\text{cm}$.
The ICP refinement search for correspondences within a $5$cm radius.

\PAR{Sequence-to-scan alignment:}
The single-image localization is only performed for keyframes, which are selected every $1$ second, $20$\degree of rotation, or $1$ meter of traveled distance.
In RANSAC, the inlier threshold depends on the detection noise and thus of the image resolution: $3\sigma$ for PnP and $1\sigma$ GPnP, as camera rigs are better constrained.
Poses with fewer than $50$ inliers are discarded.
In the rigid alignment, inliers are selected for pose errors lower than $\tau_\text{rigid}{=}(2\text{m}, 20\degree)$.
Sequences with fewer than 4 inliers are discarded.
When optimizing the pose graph, we apply the $\text{arctan}$ robust cost function to the absolute pose term, with a scale of $100$ after covariance whitening.
In the bundle adjustment, reprojection errors larger than $5\sigma$px are discarded at initialization.
A robust Huber loss function is applied to the remaining ones with a scale of $2.5$ after covariance whitening.

\PAR{Radio transfer:}
As mentioned in the main paper, currently, Apple devices only expose partial radio signals.
Notably, WiFi signals cannot be recovered, Bluetooth beacons are not exposed, and the remaining Bluetooth signals are anonymized.
This makes it impossible to match them to those recovered by other devices (e.g., HoloLens, NavVis).
To show the potential benefit of exposing these radios, we implement a simple radio transfer algorithm from HoloLens 2 devices to phones.
First, we estimate the location of each HoloLens radio detection by linearly interpolating the poses of temporally adjacent frames.
For each phone query, we aggregate all radios within at most 3m in any direction and 1.5m vertically (to avoid cross-floor transfer) with respect to the ground-truth pose.
If the same radio signal is observed multiple times in this radius, we only keep the spatially closest 5 detections.
The final RSSI estimate for each radio signal is then obtained by a distance-based weighted-average of these observations.
Note that this step is done on the raw data, after the alignment.
Thus, we can safely release radios for phone queries without divulging the ground-truth poses.

\fi

{\small
\bibliographystyle{splncs04}
\bibliography{shortstrings,references}
}

\end{document}